\documentclass{article}
\PassOptionsToPackage{numbers, compress}{natbib}
\usepackage{microtype}
\usepackage{amssymb,amsmath, amsthm}
\usepackage[pdftex]{graphicx}
\usepackage{subfigure}
\usepackage{caption}
\usepackage{wrapfig, blindtext}
\usepackage{booktabs} 
\usepackage{amssymb}
\usepackage{framed}
\usepackage{mathrsfs}
\usepackage{multirow}
\usepackage{enumerate,pifont,bigdelim,bbding}

\usepackage{hyperref}
\usepackage{bm}
\usepackage[algo2e,linesnumbered,vlined,ruled]{algorithm2e}
\usepackage{enumitem}

\newcommand{\xmark}{\ding{55}}%
\newcommand{\ip}[1] {\langle #1 \rangle }
\newcommand{\norm}[1] {\left \| #1 \right \|}
\newcommand{\inclu}[0] {\ar@{^{(}->}}

\newcommand{\RR}{\mathbb{R}}

\newcommand{\EE}{{\mathbb E}}

\newcommand{\cX}{\mathcal{X}}

\newcommand{\cO}{\mathcal{O}}

\newcommand{\cS}{\mathcal{S}}
\newcommand{\cA}{\mathcal{A}}
\newcommand{\cP}{\mathcal{P}}
\newcommand{\be}{\mathbf{e}}

\newcommand{\cB}{\mathcal{B}}
\newcommand{\cF}{\mathcal{F}}

\newcommand{\argmin}{\operatornamewithlimits{argmin}}

\newtheorem{theorem}{Theorem}[section]
\newtheorem{proposition}[theorem]{Proposition}
\newtheorem{lemma}[theorem]{Lemma}

\newtheorem{defn}[theorem]{Definition}
\newtheorem{corollary}[theorem]{Corollary}
\newtheorem{assumption}[theorem]{Assumption}

\newtheorem{example}{Example}[section]

\usepackage{mathtools}

\def\mw#1{\textcolor{red}{[MW:#1]}}

\usepackage[preprint]{neurips_2021}

\title{MARL with General Utilities via Decentralized Shadow Reward Actor-Critic}

\author{%
	Junyu Zhang\\
	Department of Electrical Engineering \\
	Center for Statistics and Machine Learning\\
	Princeton University, 
	Princeton, NJ 08544\\
	\texttt{junyuz@princeton.edu}
	\And
	Amrit Singh Bedi\\
	CISD\\
	US Army Research Laboratory \\
	Adelphi, MD 20783\\
	\texttt{amrit0714@gmail.com}
		\And
	Mengdi Wang    \\
	Department of Electrical Engineering \\
	Center for Statistics and Machine Learning\\
	Princeton University/Deepmind, 
	Princeton, NJ 08544\\
	\texttt{mengdiw@princeton.edu}
	\And	Alec Koppel\\
	CISD\\
	US Army Research Laboratory \\
	Adelphi, MD 20783\\
	\texttt{alec.e.koppel.civ@mail.mil}
}

\begin{document}
\maketitle

\begin{abstract}
We posit a new mechanism for cooperation in multi-agent reinforcement learning (MARL)
 based upon any nonlinear function of the team's long-term state-action occupancy measure, i.e.,  a \emph{general utility}. This subsumes the cumulative return but also allows one to incorporate risk-sensitivity, exploration, and priors.
We derive the {\bf D}ecentralized {\bf S}hadow Reward {\bf A}ctor-{\bf C}ritic (DSAC) in which agents alternate between policy evaluation (critic), weighted averaging with neighbors (information mixing), and local gradient updates for their policy parameters (actor). DSAC augments the classic critic step by requiring agents to (i) estimate their local occupancy measure in order to (ii) estimate the derivative of the local utility with respect to their occupancy measure, i.e., the ``shadow reward". DSAC converges to $\epsilon$-stationarity in $\mathcal{O}(1/\epsilon^{2.5})$ (Theorem \ref{theorem:final}) or faster $\mathcal{O}(1/\epsilon^{2})$ (Corollary \ref{corollary:communication}) steps with high probability, depending on the amount of communications. We further establish the non-existence of spurious stationary points for this problem, that is, DSAC finds the globally optimal policy (Corollary \ref{corollary:global}). Experiments demonstrate the merits of goals beyond the cumulative return in cooperative MARL.

\end{abstract}
\section{Introduction}\label{sec:intro}

Reinforcement learning (RL) is a framework for directly estimating the parameters of a controller through repeated interaction with the environment \cite{sutton2018reinforcement}, and has gained attention for its ability to alleviate the need for a physically exact model across a number of domains, such as robotic manipulation \cite{kober2013reinforcement}, web services \cite{zhao2018deep}, and logistics \cite{feinberg2016optimality}, and various games \cite{tesauro1994td,silver2016mastering}. In RL, an agent in a given state takes an action, and transits to another according to a Markov transition density, whereby a reward informing the merit of the action is revealed by the environment. Mathematically, this setting may be encapsulated by a Markov Decision Process (MDP) \cite{puterman2014Markov}, in which the one seeks to select the action sequence to maximize the long-term accumulation of rewards.

{In many domains, multiple agents interact in order to obtain favorable outcomes, as in finance \cite{lee2002stock}, social networks \cite{jaques2019social}, and games \cite{stone2001scaling,vinyals2019grandmaster}. In multi-agent RL (MARL) and more generally, stochastic games, a key question is the payoff structure \cite{shapley1953stochastic,bacsar1998dynamic}. We focus on common payoffs among agents, i.e., the utility of the team is the sum of local utilities \cite{busoniu2008comprehensive}, which contrasts with competitive settings where one agent's gain is another's loss, or combinations thereof \cite{littman1994markov}. }
Whereas typically {cooperative MARL} defines the global utility as the average over agents' local reward accumulations, here we define a \emph{new mechanism for cooperation} that permits agents to incorporate risk-sensitivity \cite{HuKa94,borkar2002risk,prashanth2016variance}, prior experience \cite{schaal1997learning,argall2009survey}, or exploration \cite{hazan2018provably,tarbouriech2019active}.
The usual common-payoff setting focuses on global cumulative return of rewards, which is a linear function of the the state-action occupancy measure. By contrast, the aforementioned decision-making goals define \emph{nonlinear} functions of the state-action occupancy measure \cite{Ka94I}. Such functions,
%
we call \emph{general utilities}, have recently yielded impressive performance in practice via prioritizing exploration \cite{mahajan2019maven,gupta2020uneven}, risk-sensitivity \cite{anonymous2021rmix}, and prior experience \cite{le2017coordinated,lee2019improved}. To date, however, there exists few formal guarantees for algorithms designed to optimize general utilities in multi-agent settings, to the best of our knowledge.

This gap motivates us to put forth the first decentralized MARL scheme for {general utilities}, and establish its  consistency and sample complexity. 
Our approach hinges upon first noting that the embarking point for most RL methodologies is the Policy Gradient Theorem \cite{williams1992simple,sutton2000policy} or Bellman's equation, both of which break down for general utilities. One potential path forward is a recent generalization of the PG Theorem for general utilities  \cite{zhang2020variational}, which expresses the gradient as product of the partial derivative of the utility with respect to the occupancy measure, and the occupancy measure with respect to the policy. 
However, in the team setting, this later factor is a \emph{global nonlinear function} of agents' policies, and hence does not permit decentralization. Thus, we define an agent's local occupancy measure as the joint occupancy measure of all agents' polices with all others' marginalized out, and its local general utility as any (not-necessarily concave) function of its marginal occupancy measure. The team objective, then, is the global aggregation of all local utilities. 

\begin{table}[t]
	\centering
	{
		\begin{tabular}{|l|l|l|}
			\hline
			Objective                          & Approach     & Convergence       \\ \hline
			\multirow{2}{*}{Cumulative Return} & Value-Based  \cite{kar2013cal,wai2018multi,lee2018stochastic,qu2019value,doan2019finite}  & \checkmark \multirow{5}{*}{} \rdelim\}{5}{10mm}[\parbox{20mm}{This\\Work}\hspace{-12mm}\Checkmark] \\ \cline{2-2}
			& Policy-Based \cite{zhang2018fully,chen2018communication} &         \checkmark          \\ \cline{1-2}
			Risk                               &       \cite{anonymous2021rmix}       &            \xmark       \\ \cline{1-2}
			Exploration                        &       \cite{mahajan2019maven,gupta2020uneven}       &         \xmark          \\ \cline{1-2}
			Priors                             &         \cite{le2017coordinated,lee2019improved}              &        \xmark           \\ \hline
		\end{tabular}
	}\vspace{4mm}
	\caption{ Cumulative Returns, Risk-Sensitivity, Exploration, and the incorporation of Priors are common goals in MARL, and subsumed by the general utilities considered here. We focus on the setting when agents are {\bf cooperative} and transition according to a common {\bf global} dynamics model \cite{busoniu2008comprehensive}. We defer a discussion of centralized training decentralized execution (CTDE), partial observability, and different transition models to appendices, with the understanding that our focus is on {\bf decentralized training} under {\bf full observability}. The respective technical settings of \cite{le2017coordinated,lee2019improved,mahajan2019maven,gupta2020uneven,anonymous2021rmix} are different; their inclusion here is to underscore their use of goals beyond cumulative return, which is given a conceptual underpinning for the first time in this work.
	}\label{table:intro} 
	\vspace{-0mm}
\end{table}

From this definition, we derive a new variant of the Policy Gradient [cf. \eqref{eq:policy_gradient}] where each agent estimate its policy gradient based on local information and message passing with neighbors. Specifically, we derive a model-free algorithm, {\bf D}ecentralized {\bf S}hadow Reward {\bf A}ctor-{\bf C}ritic (DSAC), that generalizes multi-agent actor-critic (see \cite{konda1999actor,konda2000actor}) beyond cumulative return \cite{zhang2018fully}. Each agent's procedure follows four stages:
(i) a marginalized occupancy measure estimation step used to evaluate the instantaneous gradient of the local utility with respect to the occupancy measure, which we dub the ``shadow reward" (shadow reward computation); 
(ii) accumulate ``shadow rewards" along a trajectory to estimate ``shadow" critic parameters (critic); 
(iii) average critic parameters with those of its neighbors (information mixing); and
(iii) a stochastic policy gradient ascent step along trajectories (actor).

\textbf{Contributions.} Overall, our contributions are:
\vspace{-0mm}
\begin{itemize}
	\item present the first MARL formulation that permits broader goals than the cumulative return and specialization among agents' roles;\vspace{-0mm}
	\item derive a variant of multi-agent actor-critic to solve this problem that employs an occupancy measure estimation step to construct the gradient of the general utility with respect to the occupancy measure, which serves as a  ``shadow reward" for the critic step;\vspace{-0mm}
	\item for $\epsilon$-stationarity with high probability, we respectively establish that DSAC requires $\mathcal{O}(1/\epsilon^{2.5})$ and $\mathcal{O}(1/\epsilon^{2})$ steps if agents exchange information once (Theorem \ref{theorem:final}) or multiple times per policy update (Corollary \ref{corollary:communication}). Under proper assumptions, we further establish the convergence to the globally optimal policy under diminishing step-sizes (Corollary \ref{corollary:global}). \vspace{-0mm}
	\item provide experimental evaluation of this scheme for exploration maximization and safe navigation in cooperative settings \cite{lowe2017multi}.\vspace{-0mm}
\end{itemize}


\section{Problem Formulation}\label{sec:problem}

Consider a Markov decision process (MDP) over the finite state space $ \cS$ and a finite action space $\cA$. For each state $s\in \cS$, a transition to state $s'\in \cS$ occurs when selecting action $a\in\cA$ according to a conditional probability distribution $s'\sim \cP(\cdot | a, s )$, for which we define the short-hand notation $P_{a}(s,s')$. Let $\xi$ be the initial state distribution of the MDP, i.e., $s_0\sim\xi$. We let $S:=|\mathcal{S}|$ denote the number of states and $A:=|\mathcal{A}|$ the number of actions. Consider policy optimization for maximizing general objectives that are nonlinear function of the {\it cumulative discounted state-action occupancy measure} under policy $\pi$, which contains the cumulative return as a special case \cite{zhang2020cautious,zhang2020variational}:
\begin{equation}\label{prob:main}
	\max_{\pi} R( \pi) :=  F(\lambda^{\pi})
\end{equation}
where $F$ is a general (not necessarily concave) functional and $\lambda^\pi$ is occupancy measure given by\vspace{-0mm}
\begin{equation}
	\label{prop:global-measure}
	\lambda^\pi(s,a) = \sum_{t=0}^\infty\gamma^t\cdot\mathbb{P}\Big(s^t = s, a^t = a\,\,\Big|\,\,\pi, s^0\sim\xi\Big)
\end{equation}
for $\forall a\in\cA, \forall s\in\cS$. For instance, often in applications one has access to demonstrations which can be used to learn a prior on the policy for ensuring baseline performance. Suppose $\bar\lambda$ is a prior state-action distribution obtained from demonstrations. One may seek to maintain baseline performance with respect to this prior via minimizing the Kullback-Liebler (KL) divergence between the normalized distribution $\hat\lambda = (1-\gamma)\lambda$ and the prior $\bar\lambda $ stated as 
$	\rho(\lambda) =  \hbox{KL}\left((1-\gamma){\lambda} || \bar\lambda \right)$.
In behavioral cloning, action information is missing, in which case one may instead consider a variant with respect to only the state occupancy measure. Other functional forms for \eqref{prob:main} are considered experimentally in Sec. \ref{sec:experiments}.

In this work, we consider the fully decentralized version of the problem in \eqref{prob:main}, where the state space $\cS$, the action space $\cA$, the policy $\pi$, and the general utility $F$ are decentralized among $N=|\mathcal{V}|$ distinct agents associated with an undirected graph $\mathcal{G}=(\mathcal{V},\mathcal{E})$ with vertex set $\mathcal{V}$ and edge set $\mathcal{E}$. Each agent $i\in\mathcal{V}$ is associated with its own local incentives and actions, detailed as follows.

\textbf{\noindent Space Decomposition}. The global state space $\cS$ is the product of $N$ local spaces $\cS_i$, i.e., $\cS = \cS_1\times\cS_2\times\cdots\times\cS_N$, meaning that for any $s\in\cS$, we may write
%
$	s = (s_{(1)}, s_{(2)},\cdots,s_{(N)})$
%
with $s_{(i)}\in\cS_i, i \in\mathcal{V}$. Each agent has access to the global state $s$, as customary of joint-action learners training in a decentralized manner under full observability \cite{kar2013cal,zhang2018fully,lee2018stochastic,wai2018multi,qu2019value,doan2019finite}. 
Similarly, the global action space $\cA$ is the product of $N$ local spaces $\cA_i$:  $\cA = \cA_1\times\cA_2\times\cdots\times\cA_N$, meaning that for any $a\in\cA$, we may write
%
$	a = (a_{(1)}, a_{(2)},\cdots,a_{(N)})$
with $a_{(i)}\in\cA_i,i \in\mathcal{V}$. Full observability means each agent $i$ has access to global actions $a$ concatenating all local ones.

\textbf{\noindent Policy Factorization}. The global policy $\pi(a|s)$ that maps global action $a$ for a given global state $s$ is defined as the product of local policies $\prod_{i=1}^N\pi^{(i)}(a_{(i)}|s)$, which prescribes statistical independence among agents' policies. For the parameterized policy $\pi_\theta(a|s)$ where $\theta\in\Theta$, we denote $\theta = (\theta_1,\theta_2,\cdots,\theta_N)$ as the parameter, so we can write
%
%
$\pi_\theta(a|s) = \prod_{i\in\mathcal{V}}\pi_{\theta_i}^{(i)}(a_{(i)}|s)$,
where the local policy of agent $i$ is parameterized by $\theta_i$. Since the global state is visible to all agents, the \emph{local policy} is based on the observation of the \emph{global state}. The parameters $\theta_i$ are kept private by agent $i$, meaning that agents must pass messages to become informed about others' incentives. 

\textbf{Local Cumulative State-Action Occupancy Measure.} Similar to the global occupancy measure $\lambda^\pi(s,a)$ [cf. \eqref{prop:global-measure}], define the \emph{local cumulative state-action occupancy measure}: 
\begin{equation}
	\label{prop:local-measure}
	\lambda^\pi_{(i)}(s_{(i)},a_{(i)}) =  \sum_{t=0}^\infty\gamma^t\cdot\mathbb{P}\Big(s^t_{(i)}= s_{(i)}, a^t_{(i)} = a_{(i)}\Big|\,\pi, s^0\sim\xi\Big)\vspace{-0mm}
\end{equation}
for $\forall a_{(i)}\in\cA_i, s_{(i)}\in\cS_i$. This local occupancy measure is the marginalization of the global occupancy measure with respect to all others' measures than agent $i$, whose indices are denoted as $\{-i\}\subset\mathcal{V}$. Via marginalization, we write
\begin{equation}
	\label{prop:local-measure-1}
	\lambda^\pi_{(i)}(s_{(i)},a_{(i)}) =  \sum_{a\in\{a_{(i)}\}\times\cA_{-i}}\sum_{s\in\{s_{(i)}\}\times\cS_{-i}}\lambda^{\pi}(s,a)
\end{equation}
with $\cA_{-i}\! \!=\! \!\Pi_{j\neq i}\cA_j$ and $\cS_{\!-i} \!\!=\!\! \Pi_{j\neq i}\cS_j$. Note that \eqref{prop:local-measure-1} is a linear transform of $\lambda^\pi$ in \eqref{prop:global-measure}.\vspace{0mm}\\
\textbf{Local Utility.}
Let $S_i=|\mathcal{S}_i|$ denote the number of local states and $A_i:=|\mathcal{A}_i|$ the number of local actions. For agent $i$, define the local utility function $F_i(\cdot):\RR^{S_iA_i}\mapsto\RR$ as a function of  $\lambda^\pi_{(i)}$, depends on $\theta_i$ when agent $i$ follows policy $\pi_{\theta_i} .$ Then, define the global utility as the sum of local ones:\vspace{-0mm}
\begin{equation}
	\label{defn:global-utility}
	R(\pi_\theta) = F(\lambda^{\pi_\theta}) := \frac{1}{N}\sum_{i=1}^N F_i\big(\lambda^{\pi_{\theta}}_{(i)}\big).\vspace{-0mm}
\end{equation}\vspace{-0mm}
Note that \eqref{defn:global-utility} is \emph{not node-separable}, and local occupancy measures \emph{depend on the global one} through \eqref{prop:local-measure-1}. This means that the policy parameters $\theta_i$ of agent $i$ depends on global policy $\pi$, and hence  on global parameter $\theta = (\theta_1,\theta_2,\cdots,\theta_N)$. This is a key point of departure from standard multi-agent optimization \cite{nedic2009distributed}.
Next we shift to deriving a variant of actor-critic that is attuned to the multi-agent setting with general utilities \eqref{defn:global-utility}.

\vspace{-0mm}
\section{Elements of MARL with General Utilities}\label{sec:algorithm}
This section develops an actor-critic type algorithm for MARL with general utilities \eqref{defn:global-utility}. One challenge is that the occupancy measure, the policy parameters, and the utility are coupled. Specifically, the value function is not additive across trajectories, and hence invalidates RL approaches tailored to maximizing cumulative returns based upon either the Policy Gradient Theorem \cite{williams1992simple,sutton2000policy} or Bellman's equation \cite{puterman2014Markov}.
To address this issue, we employ a combination of the chain rule, an additional density estimation step, and the construction of a ``shadow reward."
We first define the shadow reward and value function as follows and then will proceed towards the proposed algorithm.
\vspace{-1mm}
\subsection{Shadow Rewards and Policy Evaluation}

The general utility objective cannot be written as cumulative sum of returns. The nonlinearity invalidates the additivity, which is the origination of the definition of the conventional reward function and Q function, quantities that are central to approaches for maximizing cumulative-returns, via either dynamic programming \cite{puterman2014Markov} or policy search \cite{williams1992simple,sutton2000policy}. 
%
%
To circumvent the need for additivity, we will introduce auxiliary variables, which we call shadow rewards and shadow Q functions.
\begin{defn}[Shadow Reward and Shadow Q Function]
	\label{definition:Shadow-R-Q}
	The shadow reward $r_{\pi}:\cS\times\cA\mapsto \mathbb{R}$ of policy $\pi$ w.r.t. general utility $F$ is 
	$r^{\pi}(s,a) := \frac{\partial F(\lambda^{\pi})}{\partial \lambda(s,a)} $, with associated shadow Q function 
	$$Q^\pi_F (s,a):=\EE\big[\sum_{t=0}^{+\infty}\gamma^t\cdot r^{\pi}(s^t,a^t)\,\big|\,s^0=s, a^0=a, \pi\big].$$
\end{defn}
To understand these definitions, consider linearizing (differentiating) general utility $F$ with respect to $\lambda^\pi$. The linearized problem, via the chain rule, is equivalent to a MDP with cumulative return, with the shadow reward and Q function in place of the usual reward and Q functions:
%
\begin{equation}\label{eq:policy_gradient}
	\nabla_{\theta} F(\lambda^{\pi_{\theta}})=\EE\left[\sum_{t=0}^{+\infty}\gamma^t\cdot Q^\pi_F(s^t,a^t)\cdot\nabla_{\theta}\log\pi_{\theta}(a^t|s^t)\big|s_0\sim\xi, \pi\right].
\end{equation}
This expression for the policy gradient illuminates the centrality of the shadow reward/value function for nonlinear functions of the occupancy measure \eqref{prop:global-measure}, which motivates the generalized policy evaluation scheme we present next. 

{\bf \noindent Policy Evaluation Criterion.}
We shift to how one may compute the Shadow $Q$-function from trajectory information, upon the basis of which we can estimate the parameters of a critic. To do so, we use function approximation to parameterize the high-dimensional shadow Q-function. One simple choice is linear function approximation. That is, given a set of feature vectors $\{\phi(s,a)\in\RR^d: s\in\cS, a\in\cA\}$, we want to find some weight parameter $w\in\RR^d$ so that
\begin{equation}
	\label{defn:q-function-approx}
	Q_w(s,a) := \langle \phi(s,a) , w\rangle  \quad \forall (s,a)\in\cS\times\cA.
\end{equation}
In our algorithm, we will update a sequence of $\hat w$ to closely approximate the sequence of implicit shadow Q functions, as policy gets updated. In practice, the parametrization \eqref{defn:q-function-approx} needs not be linear. Indeed, experimentally, we consider $Q$ defined by a multi-layer neural network in Section \ref{sec:experiments}. 



Thus, the critic objective of policy $\pi$ is defined as the mean-square-error w.r.t. shadow $Q$-function:
\begin{align}
	\label{defn:critic-obj}
	\ell(w;\pi):=&\EE\bigg[\sum_{t=0}^\infty \frac{\gamma^t}{2}\left(Q_w(s^t,a^t) - Q^{\pi}_F(s^t,a^t)\right)^2\big|s^0\sim\xi, \pi\bigg]
	\nonumber
	\\
	=&  \frac{1}{2}\sum_{s,a} \lambda^{\pi}(s,a)\left(\phi(s,a){^\top} w - Q^{\pi}_F(s,a)\right){^2}.
\end{align}
Via the definition of the occupancy measure $\lambda^{\pi}$ [cf. \eqref{prop:global-measure}], the expectation may be substituted by weighting factors in the summand on the second line.
%
We assume features $\{\phi(s,a)\}_{s\in\cS,a\in\cA}$ are bounded, as is formalized in Sec. \ref{sec:convergence}. With the shadow reward and associated $Q$-function (Definition \ref{definition:Shadow-R-Q}), the policy evaluation criterion \eqref{defn:critic-obj}, and its smoothness properties with respect to critic parameters $w$ in place (Sec.  \ref{sec:convergence}), we expand on their role in the multi-agent setting.

\begin{algorithm2e}[t]
	\caption{{\bf D}ecentralized {\bf S}hadow Reward {\bf A}ctor-{\bf C}ritic (DSAC)}
	\label{alg:stochastic-batch-2ts}
	\textbf{Input:} initial policy $\theta^0$; actor step-sizes $\{\eta_\theta^k\}$; Batch sizes $\{B_k\}$; Episode lengths $\{H_k\}$; initial critic $W^0:=[w^0_1,w^0_2,...,w^0_N]\in\RR^{d}$ with $w^0_i=w^0_j$, $\forall i,j$; critic step-size $\{\eta_w^k\}$; mixing matrix $M\in\RR_+^{N\times N}$; mixing round $m\geq1.$\\ 
	\For{ iteration $k=0,1,2,...$}{
		Perform $B_k$ Monte Carlo rollouts to obtain trajectories $\tau = \{s^0,a^0,\cdots,s^{H_k},a^{H_k}\}$ with initial dist. $\xi$, policy $\pi_{\theta^k}$ collected as batch $\cB_k$. \\  
		\For{agent $i = 1,2,...,N$}{
			Compute empirical local  occupancy  measure 
			\begin{equation}
				\label{eq:occupancy_measure_estimator}
				\hat\lambda^k_{i} = \frac{1}{B_k}\sum_{\tau\in\cB_k} \sum_{t=0}^{H_k}\gamma^t \cdot {\bf e}\left(s^{t}_{(i)}, a^{t}_{(i)}\right).
			\end{equation}
			Estimate shadow reward $\hat r^k_{i} = \nabla_{\lambda_{i}} F_{i}(\hat\lambda^k_{i})$.}
		\For{agent $i=1,2,...,N$}{
			With localized policy gradient estimate $G_{w_i}(\tau,r_i,w_i) = \sum_{t=0}^{H}\gamma^t\cdot(Q_{w_i}(s^{t},a^{t})-\hat Q^{t}_{i})\cdot\nabla_{w_i}Q_{w_i}(s^{t},a^{t})$, compute \vspace{-0mm}
			$$\widehat{\Delta}_{w_{i}}^k = \frac{1}{B_k}\sum_{\tau\in\cB_k} G_{w_{i}}(\tau,\hat r^k_{i},w_{i}^k)\;, \qquad w_{i}^{k+1} = w_{i}^k - \eta_w^k\widehat{\Delta}_{w_{i}}^k.\vspace{-4mm}$$
		}	
		\For{$iter = 1,...,m$}{
			\For{agent $i=1,2,...,N$}{
				Exchange information with neighbours: $w_{i}^{k+1} = \sum_{\{j:(j,i)\in\mathcal{E}\}} M(j,i)\cdot w_i^{k+1}.$\vspace{-0mm}}}
			\For{agent $i=1,2,...,N$}{
		With $G_{\theta_i}(\tau,w_i) = \sum_{t=0}^{H}\gamma^t Q_{w_i}(s^{t},a^{t})\nabla_{\theta_i} \log \pi_{\theta_i}^{(i)}(a^{t}_{(i)}|s^{t})$, update the policy:\vspace{-0mm}
		$$\widehat{\Delta}^k_{\theta_{i}}:= \frac{1}{B_k}\sum_{\tau\in\cB_k} G_{\theta_{i}}(\tau,w_{i}^{k+1})\;, \qquad\theta_{i}^{k+1} =  \theta_{i}^k + \eta_\theta^k \widehat{\Delta}_{\theta_{i}}^k.$$
	}
	}
\end{algorithm2e}\vspace{-0cm}
%

\subsection{Multi-Agent Optimization for Critic Estimation}
Setting aside the issue of policy parameter updates for now, we focus on estimating the global general utility. The shadow Q-function and shadow reward (Definition \ref{definition:Shadow-R-Q}) depend on global knowledge of all local utilities, which are unavailable as local incentives are local only. To mitigate this issue, we introduce their localized components, which together comprise the global shadow Q-function and reward. Specifically, define the local shadow reward $r_{i}^{\pi}$ for agent $i$:
\begin{equation}
	\label{defn:reward-multi-agent-local}
	r_{i}^{\pi}(s_{(i)},a_{(i)}) := \frac{\partial F_i(\lambda_{(i)}^{\pi})}{\partial \lambda_{(i)}(s_{(i)},a_{(i)})} \;, \forall (s_{(i)},a_{(i)})\in\cS_i\times\cA_i.
\end{equation}
Clearly, it holds that $r^{\pi}(s,a) = \frac{1}{N}\sum_{i=1}^Nr_{i}^{\pi}(s_{(i)},a_{(i)}).$
Based on the local observation of the its own shadow reward, agent $i$ may access its local shadow Q-function $Q: \cS \times \cA \rightarrow \mathbb{R}$: 
\begin{equation}
	\label{defn:Q-multi-agent-local}
	Q^{\pi}_{i}(s,a):=\EE\left[\sum_{t=0}^{+\infty}\gamma^t\cdot r_{i}^{\pi}\big(s^t_{(i)},a^t_{(i)}\big)\,\big|\,s^0=s, a^0=a, \pi\right],
\end{equation} 
for $\forall (s,a)\in\cS\times\cA$. Therefore, we also have $Q^{\pi}_F(s,a) = \frac{1}{N}\sum_{i=1}^NQ^{\pi}_{i}(s,a)$.  
%
Then, each agent $i$ seeks to estimate common critic parameters $w$ that well-represent its shadow $Q$ function in the sense of minimizing the global mean-square error \eqref{defn:critic-obj}. By exploiting the aforementioned node-separability and introducing a localized critic parameter vector $w_i$ associated to agent $i$, this may equivalently be expressed as a consensus optimization problem \cite{nedic2009distributed}:
%
%
\begin{align}
	\label{defn:critic-obj-marl-1}
	\min_{\{w_i\}_{i=1}^N}& \frac{1}{N}\sum_{i=1}^N\ell_i(w_i;\pi) 
	\ \ \mbox{s.t.} \ \  w_i = w_j, (i,j)\in\mathcal{E}\; ,\   
	\nonumber
	\\
	&\ell_i(w_i;\pi) := \EE\left[\sum_{t=0}^\infty \frac{\gamma^t}{2}\left(Q_{w_i}(s^t,a^t) - Q^{\pi}_{F_i}(s^t,a^t)\right)^{2}\big|s^0\sim\xi, \pi\right]. 
	%
\end{align}
where the local policy evaluation criterion is defined as 
%
This formulation allows agent $i$ to evaluate its policy with respect to global utility \eqref{defn:global-utility} through the local criterion $ \ell_i(w_i;\pi)$ as a surrogate for that which aggregates global information \eqref{defn:critic-obj}, when consensus over local parameters $w_i$ is imposed. Next, we incorporate  solutions to \eqref{defn:critic-obj-marl-1} into the critic step together with a policy parameter $\theta_i$ update along stochastic ascent directions via \eqref{eq:policy_gradient} for the actor to assemble DSAC.

%
%

\subsection{Decentralized Shadow Reward Actor-Critic}\label{sec:alg_actorcritic}

Next, we put together these pieces to present  {\bf D}ecentralized {\bf S}hadow Reward {\bf A}ctor-{\bf C}ritic (DSAC) as Algorithm \ref{alg:stochastic-batch-2ts} (see Fig. \ref{algorithm_execution} in the appendix for the flow diagram). This scheme allows agents to keep their local utilities $F_i$, and policies $\pi_{\theta_i}$ with associated parameters $\theta_i$ private. The agents share a common function approximator for the shadow Q function. Further, they retain local copies $w_i$ of the shadow critic parameters, which they communicate to neighbors according to the network structure defined by edge set $\mathcal{E}$ and mixing matrix $M$ to be subsequently specified. 
%
%
Algorithm \ref{alg:stochastic-batch-2ts} proceeds in four stages: (i) density estimation step for to obtain the shadow reward; (ii) shadow critic updates; (iii) information mixing via weighted averaging; and (iv) actor updates. 
Each step is detailed in Algorithm \ref{alg:stochastic-batch-2ts}, with step-by-step instructions in Appendix \ref{apx:guide}.


\section{Consistency and Sample Complexity} \label{sec:convergence}
In this section, we study the finite sample performance of Algorithm \ref{alg:stochastic-batch-2ts}. 
We show $\tilde{\mathcal{O}}(\epsilon^{-2.5})$ (Theorem \ref{theorem:final}) or $\tilde{\mathcal{O}}(\epsilon^{-2})$ (Corollary \ref{corollary:communication}) sample complexities to obtain $\epsilon$-stationary points of global utility, depending on the number of communications per step, akin to best known rates for non-concave expected maximization problems \cite{shapiro2014lectures}. We also establish the nonexistance of spurious extrema for this setting, indicating the convergence to global optimality (Corollary \ref{corollary:global}). 
%
%
%
%
%
Before continuing, we present a few key technical conditions
for the utility $F$, the policy $\pi_{\theta}$, the mixing matrix $M$, and the critic approximation. 
\begin{assumption}
	\label{assumption:Utility}
	For utility $F$ [cf. \eqref{defn:global-utility}], we assume:	\\
	{\bf(i).} $F_i(\cdot)$ is private to agent $i$, for $\forall i$.\\
	{\bf(ii).} $\exists C_{F}>0$ s.t. $\|\nabla_{\lambda_{(i)}}  F_i(\lambda_{(i)})\|_\infty\leq C_F$ in a neighbourhood of the occupancy measure set, $\forall i$.\\
	{\bf(iii).}  $\exists L_\lambda>0$ s.t. $\|\nabla_{\lambda_{(i)}} F_i(\lambda_{(i)})-\nabla_{\lambda_{(i)}} F_i(\lambda'_{(i)})\|_\infty\leq L_\lambda\|\lambda_{(i)}-\lambda'_{(i)}\|$, for $\forall i$.\\
	{\bf(iv).} $\exists L_\theta>0$ s.t.   $F\circ\lambda(\cdot)$ is $L_\theta$-smooth.
\end{assumption} 
\begin{assumption}
	\label{assumption:parameterization}
	For $\pi_{\theta}$ and the occupancy measure $\lambda^{\pi_{\theta}}$, we assume: \\
	{\bf(i).} The local policy $\pi_{\theta_{i}}^{(i)}$ is private to each agent $i$.\\
	{\bf(ii).}  $\exists$ $C_{\pi}>0$ s.t. for any agent $i$, the score function is upper bounded: $\|\nabla_{\theta_i} \log \pi_{\theta_i}^{(i)}(a_{(i)}|s)\|\leq C_{\pi}$, for $\forall$ $\theta$ and $\forall (s,a)$. \\
	{\bf(iii).} $\exists$ $\ell_\theta>0$ s.t. $\|\lambda^{\pi_\theta}-\lambda^{\pi_{\theta'}}\|\leq \ell_\theta\|\theta-\theta'\|.$ 
\end{assumption}\vspace{-0mm}
\begin{assumption}
	\label{assumption:mixing-matrix}
	The mixing matrix $M$ is a doubly stochastic matrix satisfying: \\
	{\bf(i).} $M\in\mathbb{S}_+^{N\times N}$, $M\!(i,j)>0$ iff. $(i,j)\in\mathcal{E}$.\\
	{\bf(ii).} $M\cdot\mathbf{1}_N = \mathbf{1}_N$, where $\mathbf{1}_N\in\RR^N$ is an all-ones vector.\\
	{\bf(iii).} Let the eigenvalues of $M$ be $1 = \sigma_1(\!M)\!>\!\sigma_2(\!M)\!\geq\!\cdots\!\geq\!\sigma_N(\!M)$. We define $\rho\!:=\!\max\{|\sigma_2(\!M)|,|\sigma_N(\!M)|\}\!<\!1$.
\end{assumption}
\begin{assumption}
	\label{assumption:model-err}
	For $\forall \theta$, define the optimal critic parameter  {$w^*(\theta): =\argmin_w \frac{1}{N}\sum_{i=1}^{N}\ell_i(w;\pi_{\theta})$}. We assume that $\exists W>0$ s.t. {$E^2_\theta= \sum_{i=1}^N\left\|\nabla_{\theta_i} F(\lambda^{\pi_{\theta}})-\Delta_{\theta_i}\right\|^2 \leq W$}, for $\forall \theta$, where 
	$$\Delta_{\theta_{i}} := \EE\big[\sum_{t=0}^{+\infty}\gamma^t\cdot Q_{w^*(\theta)}(s^t,a^t)\cdot\nabla_{\theta_i}\log\pi_{\theta_i}^{(i)}(a^t|s^t)\big| s^0\sim \xi, \pi_{\theta}\big]$$ is the PG estimate under $w^*(\theta)$. 
\end{assumption}
Assumption \ref{assumption:Utility} requires the boundedness and Lipschitz continuity of the gradient of the utility function.  Assumption \ref{assumption:parameterization} ensures that the score function is bounded, and the occupancy measure is Lipschitz w.r.t. the policy parameters. These conditions are common to RL algorithms focusing on occupancy measures in recent years \cite{hazan2018provably,zhang2020variational}, and are automatically satisfied by common policies such as the softmax. Assumption \ref{assumption:mixing-matrix} holds for any undirected connected loop-free static graph \cite{chung1997spectral}. Assumption \ref{assumption:model-err} states that the feature mis-specification error is uniformally upper bounded by $W$. Besides, we also make the following assumptions. 

For the shadow Q function and the occupancy measure, we make the following assumption. This assumption can be implied by more basic assumptions on the boundedness and Lipschitz continuity of the score functions. We prefer directly assuming the Lipschitz continuity of the shadow Q function in order to avoid the heavy notations.
\begin{assumption}
	\label{assumption:Q-occupancy}
	$\exists\ell_Q,\ell_\lambda>0$ s.t. for $\forall(s,a)\in\cS\times\cA$, $\forall \theta,\theta'$, it holds that  $|Q_F^{\pi_{\theta}}(s,a)-Q_F^{\pi_{\theta}}(s,a)|\leq \ell_Q\|\theta - \theta'\|$, and $|\lambda^{\pi_{\theta}}(s,a)-\lambda^{\pi_{\theta}}(s,a)|\leq\ell_\lambda\|\theta - \theta'\|.$
\end{assumption}
We also assume the boundedness of the feature vectors.
\begin{assumption}
	\label{assumption:critc-features}
	$\exists C_\phi>0$ s.t. $\|\phi(s,a)\|\leq C_\phi$, $\forall (s,a)$.
\end{assumption}
As a consequence of the boundedness of the features, the critic objective function has Lipschitz continuous gradients.
\begin{proposition}\label{prop:lipschitz}
	Regardless of policy $\pi_{\theta}$, critic objective $\ell(w;\pi_{\theta})$ [cf. \eqref{defn:critic-obj}] is  $L_w:= \frac{C_\phi^2}{1-\gamma}$ smooth.
\end{proposition}
This can be shown by directly computing the Hessian matrix of the critic objective function as $\nabla_w^2\ell(w;\pi_{\theta}) = \sum_{s,a}\lambda^{\pi_{\theta}}(s,a)\!\cdot\!\phi(s,a)\phi(s,a)^\top.$
Consequently, $L_w \leq \|\nabla_w^2\ell(w;\pi_{\theta})\|_F\leq \frac{C_\phi^2}{1-\gamma}.$ 

Throughout the iterations of Algorithm \ref{alg:stochastic-batch-2ts}, we also make the following assumption on the critic objective function. 
\begin{assumption}
	\label{assumption:critic-SC} $\ell(w;\pi_{\theta^{k}})$ is $\mu_w$-strongly convex for all $k$. 
\end{assumption}
%
%
%

Assumption \ref{assumption:critic-SC} means that the minimum eigenvalue of the feature covariance matrix  $\sum_{s,a}\lambda^{\pi_{\theta^{k}}}(s,a)\cdot\phi(s,a)\phi(s,a)^\top$ is uniformly lower bounded by some constant $\mu_w>0$. Note that the shadow reward $r^\pi$ is changing with iteration index $k$, and consequently we cannot assume that the fitted shadow Q-function perfectly tracks the true shadow Q-function. Motivated by the subtleties of the quality of a feature representation, we further place a condition on the shadow value function approximation error.

Next, we present the proof of the finite sample performance of the algorithm with details provided in the appendices. Based on the smoothness of of the utility function $F(\lambda^{\pi_{\theta}})$, the standard Taylor's expansion allows us to write:
\begin{lemma}
	\label{lemma:sufficient-ascent}
	For Algorithm \ref{alg:stochastic-batch-2ts}, if the step size of $\theta$ satisfies $\eta_\theta^k\leq1/4L_\theta$, then 
	\begin{eqnarray} 
		\label{eqn:descent-sto-1}
		F(\lambda^{\pi_{\theta^{k+1}}}) - F(\lambda^{\pi_{\theta^{k}}})\!\geq\!  \frac{\eta_\theta^k}{4}\left\|\nabla_{\theta}F(\lambda^{\pi_{\theta^{k}}})\right\|^2- \frac{3\eta_\theta^k}{4} \sum_{i=1}^{N}\left\|\nabla_{{\theta_i}}F(\lambda^{\pi_{\theta^{k}}})-\widehat{\Delta}_{\theta_i}^k\right\|^2.
	\end{eqnarray} 
\end{lemma}
Not surprisingly, the key here is bounding the gradient estimation error term $\sum_{i=1}^{N}\left\|\nabla_{{\theta_i}}F(\lambda^{\pi_{\theta^{k}}})-\widehat{\Delta}_{\theta_i}^k\right\|^2$, which is caused jointly by the stochastic sampling error, the shadow Q-function approximation error, and multi-agent concensus error, as is characterized below.
\begin{lemma}
	\label{lemma:stochastic-grad-err}
	Let $\delta_k\in(0,1)$ be some small failure probability. Then following inequalities hold. 
	\begin{enumerate}[label=(\roman*)]
		\item \label{lemma:stochastic-grad-err_1} For the ease of notation, define the error matrix of the critic estimators as 
		$$\zeta_W^k: = [\nabla_{w_1}\ell_1(w_1^k;\pi_{\theta^{k}}) -\widehat{\Delta}_{w_1}^k; \cdots ; \nabla_{w_N}\ell_N(w_N^k;\pi_{\theta^{k}}) -\widehat{\Delta}_{w_N}^k].$$ 
		For the critic gradient estimator, we have 
		\begin{eqnarray}
			\label{lm:stochastic-grad-err-3}
			\mathbf{Prob}\left(\left\|\zeta_W^k\right\|^2_F\geq\mathcal{E}_{w}^{k}\right) \leq 2N\delta_k.
		\end{eqnarray}
		where $\mathcal{E}_w^k = \cO\Big( \frac{\log(1/\delta_k)C_\phi^2}{(1-\gamma)^2B_k}\Big(C_\phi^2\sum_{i=1}^N\|w_i^k\|^2 + \frac{NC_F^2}{(1-\gamma)^2}+NL_\lambda^2\Big) +  \gamma^{2H_k}\Big)$.
		\item \label{lemma:stochastic-grad-err_2} Denote $w_*^{k+1} = w^*(\theta^k)$ (see Assumption \ref{assumption:model-err}) as the ideally fitted critic parameter. For the actor gradient estimator, there is a positive random variable $\zeta_\theta^k$ (defined in \eqref{eq:zeta_theta}) s.t.
		\begin{eqnarray}
			\label{lm:stochastic-grad-err-4}
			\sum_{i=1}^{N}\!\|\widehat{\Delta}_{\theta_i}^k \!-\! \nabla_{\theta_i} \!F(\lambda^{\pi_{\theta^{k}}})\|^2\leq \frac{3C_\phi^2C_\pi^2}{(1\!-\!\gamma)^2}\sum_{i=1}^{N}\|w_i^{k\!+\!1}\!-\!w_*^{k\!+\!1}\|^2 \! + \!\zeta_\theta^k\!+\! 6E_{\theta^{k}}^2\!+\!\cO(\gamma^{2H_k})
		\end{eqnarray}
		where $\mathbf{Prob}\left( \zeta_\theta^k \geq  \mathcal{E}_{\theta}^k\right)\leq N\delta_k$, $\mathcal{E}_{\theta}^k = \cO\Big(\frac{C_\phi^2C_\pi^2}{(1-\gamma)^2}\cdot \frac{N\log(1/\delta_k)\|w_*^{k\!+\!1}\|^2}{B_k}\Big).$
	\end{enumerate}
\end{lemma}
See Appendix \ref{appendix_taylor} for proof.
The inequality \eqref{lm:stochastic-grad-err-4} characterizes the error of the gradient estimators. However, in this inequality, the bound on the term $\sum_{i=1}^N\|w_i^{k+1}-w_*^{k+1}\|^2$ needs further study.  This error exists because Algorithm \ref{alg:stochastic-batch-2ts} makes only one stochastic gradient update. Though minimizing the critic objective function exactly in each iteration eliminates this term, it is often sample inefficient to do so in practice. By splitting the error into two parts: $\sum_{i=1}^N\|w_i^{k+1}-w_*^{k+1}\|^2 \leq 2\sum_{i=1}^N\|w_i^{k+1}-\bar w^{k+1}\|^2 + 2N\|\bar w^{k+1}-w_*^{k+1}\|^2$, where $\bar w^{k+1} = \frac{1}{N}\sum_{i=1}w_i^{k+1}$, we observe that the first part is a consensus error, which we study in Lemma \ref{lemma:marl-dis-from-avg-sto}. The second component is the optimality gap of the critic fitting problem which is the focus of Lemma \ref{lemma:marl-contraction-sto}. 
\begin{lemma}
	\label{lemma:marl-dis-from-avg-sto}
	Let the sequence $\{w_i^k\}$ be generated by Algorithm \ref{alg:stochastic-batch-2ts}. Define $\bar w^k:=\frac{1}{N}\sum_{i=1}^Nw_i^k.$ Then as long as the step size $\eta_w^k\leq 1/L_w$, it holds that 
	\begin{eqnarray}
		\sum_{i=1}^N \!\|w^{k}_i\!-\!\bar w^{k}\|^2 \!\leq\! 2\bigg(\!\max_{k'\leq k}\!\|\zeta_W^{k'}\!\|_F^2 \!+\! \frac{NC_F^2C_\phi^2}{(1-\gamma)^4}\bigg)\!\cdot\!\bigg(\sum_{k'=0}^k\eta_w^{k'}\rho^{m(k-k')}\bigg)^2\!\cdot\!\rho^{2m}.
	\end{eqnarray}
\end{lemma} 
\begin{lemma}
	\label{lemma:marl-contraction-sto}
	There $\exists C_w>0$ s.t. $\|w^{k+1}_*-w^k_{k}\|\leq C_w\|\theta^k-\theta^{k-1}\|$ for all the iterates. If the step sizes satisfy $\eta_\theta^{k-1}\leq\frac{(1-\gamma)\mu_w\eta_w^k}{4\sqrt{3N}C_wC_\phi C_\pi}$, then 
	\begin{align}
		&\|\bar w^{k+1}-w^{k+1}_*\|^2  
		\nonumber
		\\
		& \leq   \left(1-\frac{\eta_w^k\mu_w}{4} \right)\cdot\|\bar w^k - w^{k}_* \|^2 + \frac{2C_w^2(\eta_\theta^{k-1})^2}{\eta_w^k\mu_w}\|\nabla_{{\theta}} F(\lambda^{\pi_{\theta^{k-1}}})\|^2 + \frac{2\eta_w^k\|\zeta_W^k\|_F^2}{\mu_w\cdot N}\nonumber
		\\
		&\qquad + \frac{2C_w^2(\eta_\theta^{k-1})^2}{\eta_w^k\mu_w}\left(\zeta_\theta^{k-1} +6E_{\theta^{k-1}}^2+\cO( \gamma^{2H_{k-1}}) + \frac{6C_\phi^2C_\pi^2}{(1-\gamma)^2}\cdot\sum_{i=1}^{N}\| w^{k}_i-\bar w^{k}\|^2\right).
	\end{align}
\end{lemma}
See proof in Appendix \ref{appdx:lm-marl-contraction-sto}. Specifically, the existence of such a constant $C_w$ is proved in Appendix \ref{appdx:support}.


Next, we construct the following potential function with a carefully selected constant $\alpha$
%
\begin{eqnarray}
	\label{defn:potential}
	R_{k}:= F(\lambda^{\pi_{\theta^{k}}}) - \alpha\|\bar w^k-w^k_*\|^2.
\end{eqnarray}
Taking the advantage of the contraction property of {$\|\bar w^{k}\!-\!w^{k}_*\|^2$} and this specific potential function, as well as the fact that $\max_{k\geq 0}\{\|w_*^{k+1}\|,\|w_i^k\|\}\leq D_w$ for some constant $D_w>0$ (proved in Appendix \ref{appdx:support}),we characterize algorithm performance in terms of optimization error, the feature mis-specification error, the stochastic PG approximation error, and the multi-agent consensus error. 
\begin{lemma} 
	\label{lemma:cvg-sto-2ts}
	For Algorithm \ref{alg:stochastic-batch-2ts}, if for any $k\geq0$ we choose the step sizes to be  $\eta_w^{k+1}\leq\eta_w^{k}\leq 1/L_w$, and 
	$\eta_\theta^{k} =  \min\left\{\frac{(1-\gamma)\mu_w\eta_w^{k+1}}{C_wC_\phi C_\pi}\cdot\frac{1}{\max\{4\sqrt{3N},6\sqrt{10}\}}, \frac{1}{4L_\theta}\right\},$ and we set $\alpha =\frac{18C_\phi^2C_\pi^2}{(1-\gamma)^2\mu_w}\cdot\max_{k\geq 0}\{{\eta_\theta^k}/{\eta_w^k}\}$ in the potential function \eqref{defn:potential}, then with probability $1-3N\sum_{k=0}^T\delta_k$, we have 
	\begin{align} 
	\frac{\sum_{k=1}^T\eta_\theta^k\left\|\nabla_{\theta}F(\lambda^{\pi_{\theta^{k}}})\right\|^2}{\sum_{k=1}^T\eta_\theta^k} 
		\leq& \cO\left(\frac{R_{T+1} - R_{1}}{\sum_{k=1}^{T}\eta_\theta^k}   \right) +\cO\left(\frac{\sum_{k=0}^{T}\eta_\theta^kE_{\theta^{k}}^2}{\sum_{k=1}^{T}\eta_\theta^k}   \right) \nonumber\\
		&+ \cO\left(\frac{\sum_{k=1}^{T}\eta_\theta^{k-1}\cdot\big(\frac{N\log(1/\delta_k)}{B_k}+\gamma^{2H_k}\big)}{\sum_{k=1}^{T}\eta_\theta^k}\right)\nonumber\\
		&\quad+ \cO\left(\frac{\sum_{k=1}^T\eta_\theta^k\cdot\left(\sum_{k'=0}^k\eta_w^{k'}\rho^{k-k'}\right)^2}{\sum_{k=1}^{T}\eta_\theta^k}\cdot\rho^{2m}\right),
	\end{align}
	where the four terms stand for the optimization error, the feature mis-specification error, the stochastic PG approximation error, and the multi-agent consensus error. 
\end{lemma}
Upon the basis of Lemma \ref{lemma:cvg-sto-2ts}, we select parameters to conclude the statements in Theorem \ref{theorem:final}, which is expanded upon at length in Appendix \ref{appendix_theorem}.

Combining the above steps and suitably specify the parameters, we have the final theorem. \vspace{0mm}


\begin{theorem}
	\label{theorem:final}
	Under Assumption \ref{assumption:critc-features}, \ref{assumption:Utility}, \ref{assumption:parameterization}, \ref{assumption:critic-SC} and \ref{assumption:mixing-matrix}, with one communication round per iteration, i.e. $m=1$, Algorithm \ref{alg:stochastic-batch-2ts} satisfies, under the following parameter selections: \\ 
	(i) For final iteration $T \!=\! \cO(\epsilon^{-1.5})$, trajectory lengths $H_k \equiv \cO\left({\log(1/\epsilon)}/{1-\gamma}\right)$, $\delta_k \!\equiv \!{\delta}/{(3N(T\!+\!1))}$, $\delta\in(0,1)$, batch sizes $B_k \equiv \log(1/\delta_k)\epsilon^{-1}$, constant step-sizes $\eta_w = \cO(\sqrt{\epsilon})$, {$ \eta_\theta=\min\left\{\frac{(1-\gamma)\mu_w\eta_w}{C_wC_\phi C_\pi}\!\cdot\!\frac{1}{\max\{4\sqrt{3N},6\sqrt{10}\}}, \frac{1}{4L_\theta}\right\}\!=\! \cO(\sqrt{\epsilon}),$} then \vspace{-0mm}
	\begin{equation*} 
		\frac{1}{T}\sum_{k=1}^T\left\|\nabla_{\theta}F(\lambda^{\pi_{\theta^{k}}})\right\|^2
		\leq \cO\left(\epsilon + W \right). \quad w.p.\quad 1-\delta
	\end{equation*}
	(ii)  For unspecified final iteration $T$, we adaptively set: $\delta_k =  \frac{2\delta}{N\pi^2(k+1)^2}$, $\delta\in(0,1)$, trajectory lengths $H_k = \cO((1-\gamma)^{-1}\log(k+1))$,  batchsizes $B_k = \log(1/\delta_k)(k+1)^{\frac{2}{3}}$, and step-sizes {$\eta_\theta^{k} =  \min\!\Big\{\!\frac{(1\!-\!\gamma)\mu_w\eta_w^{k\!+\!1}}{C_wC_\phi C_\pi}\!\cdot\!\frac{1}{\max\{4\sqrt{3N},6\sqrt{10}\}}, \frac{1}{4L_\theta}\Big\},$ $\eta_w^k = \min\{(k+1)^{-\frac{1}{3}},L_w^{-1}\}$},  then 
	\begin{equation*} 
		\frac{\sum_{k=1}^T\eta_\theta^k\big\|\nabla_{\theta}F(\lambda^{\pi_{\theta^{k}}})\big\|^2}{\sum_{k=1}^T\eta_\theta^k}  \!\leq\! \cO\!\left(\frac{\log T}{T^{\frac{2}{3}}}+W\right), \ \ w.p. \!\quad 1-\delta
	\end{equation*}
	%
	In either case, Algorithm \ref{alg:stochastic-batch-2ts} requires $\tilde{\cO}(\epsilon^{2.5})$ samples to satisfy {$\frac{\sum_{k=1}^T\eta_\theta^k\left\|\nabla_{\theta}F(\lambda^{\pi_{\theta^{k}}})\right\|^2}{\sum_{k=1}^T\eta_\theta^k}\leq \cO(\epsilon+W)$.}
\end{theorem}
Next, we establish that for concave general utilities \eqref{prob:main}, there are no spurious stationary points.
%
%
\begin{corollary}[Convergence to global optimality]
	\label{corollary:global}
	Suppose  $F$ is concave, and the shadow Q function $Q_F$ is realizable, i.e., $W= 0$ in Assumption \ref{assumption:model-err}. For $\pi_{\theta}$ satisfying Assumption 1 of \cite{zhang2020variational}, every stationary point is a global optimizer. In Theorem \ref{theorem:final}(ii), if we further let $\bar\theta_T$ be the parameter randomly chosen from $\{\theta^k\}_{k=1}^T$ where $\bar\theta_T= \theta^k$ w.p. $\eta_\theta^k/(\sum_{k'=1}^T\eta_\theta^{k'})$, then $\lim_{T\to\infty}\EE[\|\nabla_{\theta}F(\lambda^{\pi_{\bar\theta_T}})\|^2] = 0$ w.p. $1-\delta$. Thus, Algorithm \ref{alg:stochastic-batch-2ts} converges to the set of global optimizers.
\end{corollary}
Next we spotlight the role of the number of communication steps in the convergence rate.
\begin{corollary}[Multiple-round communication]
	\label{corollary:communication}
	Suppose multiple-round communication is allowed, i.e., $m>1$. Under the same parameter selections as Theorem \ref{theorem:final}(i), while setting final iteration index $T = \epsilon^{-1}$, communication rounds $m = \cO((1-\rho)^{-1}\log(\epsilon^{-1}))$, and the step-sizes  {$\eta_\theta^{k} \equiv\min\Big\{\frac{(1-\gamma)\mu_w/L_w}{C_wC_\phi C_\pi}\cdot\frac{1}{\max\{4\sqrt{3N},6\sqrt{10}\}}, \frac{1}{4L_\theta}\Big\}$, $\eta_w^k\equiv L_w^{-1}$}, then the total sample complexity is $\cO(\epsilon^{-2})$.\vspace{-0mm}
\end{corollary}
Namely, with additional communication rounds $m \!=\!\cO(\!(1\!-\!\rho)^{-1}\!\log(\epsilon^{-1})\!)$ per iteration, the convergence rate refines from $\cO(\epsilon^{-2.5})$ to $\cO(\epsilon^{-2})$.
%
Next, we investigate the experimental merit of the proposed approach for giving rise to emergent teamwork among multiple agents across various tasks.

\section{Experimental Results}\label{sec:experiments}
We experimentally investigate the merit of Algorithm \ref{alg:stochastic-batch-2ts} in the context of both single and multi-agent problems. The single-node case ($N=1$)  bears investigation as the proposed scheme is a new way to solve RL problems with general utilities relative to \cite{zhang2020variational}. For this case, we consider the continuous \emph{MountainCar} environment of OpenAI Gym \cite{brockman2016openai}. The additional experiments for single agent settings to Appendix  \ref{additional_single}.
%

\subsection{Concept of Shadow Reward}
\begin{figure}[ht]
	\vspace{-0mm}
	\centering
	\subfigure[ Shadow reward ]{\includegraphics[scale=0.19]{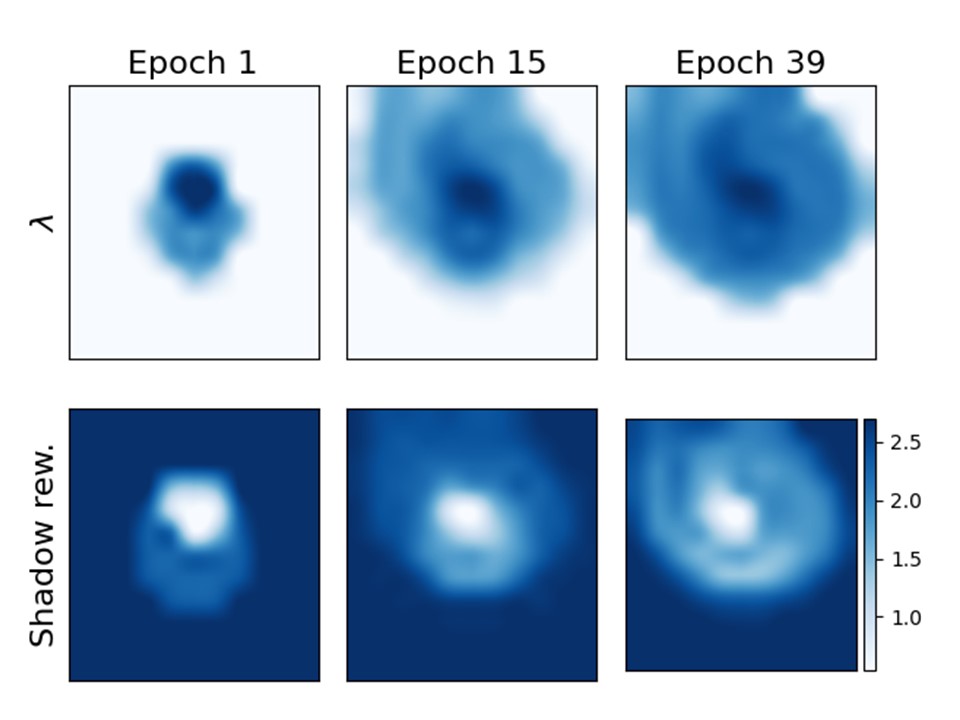}\label{fig1_entropy11}}\hspace{-3mm}
	\subfigure[ Entropy comparison]{\includegraphics[scale=0.30]{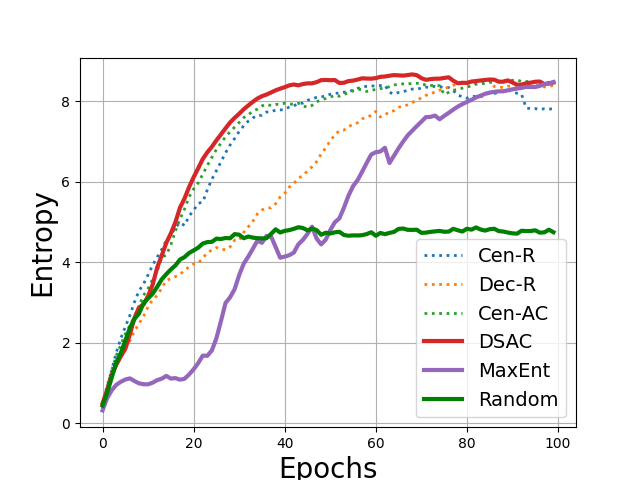}\label{fig_entropy_cost}}
\hspace{-3mm}
	\subfigure[ \vspace{-0mm}State space coverage]{\includegraphics[scale=0.28]{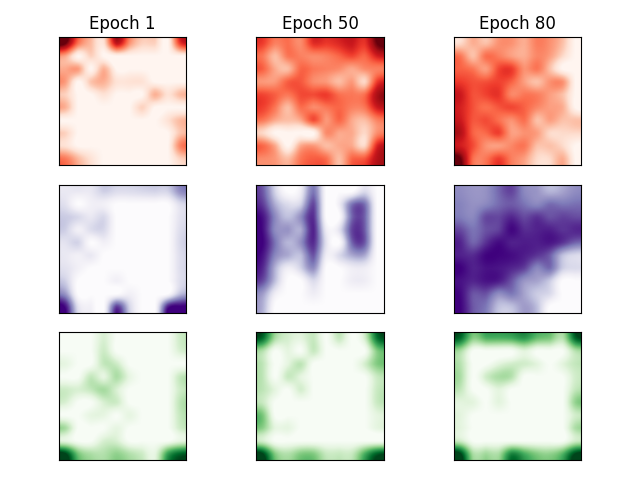}\label{fig_entropy_occupancy}}
\hspace{-3mm}	\captionof{figure}{ (a) Occupancy measure {(first row)}  and shadow reward {(second row)} for \emph{MountainCar} environment. {Each subplot represents a heatmap for two dimentional state space.} Observe that over the course of training the measure and shadow reward's coverage of the state and action spaces grows, as a consequence of selecting actions towards maximizing the entropy of the occupancy measure. (b) Entropy comparisons for exploration maximization in a \emph{cooperative multiagent} environment, (c) Agent 1 marginalized occupancy measure. For the DSCA implementation, each agents needs to estimate only $100$ dimensional marginalized occupancy measure while for the centralized counterpart MaxEnt, we need to estimate $10^4$ dimensional occupancy measure making it slow in practice. }
	\label{fig_multi_entropy}\vspace{-0mm}
\end{figure}

To understand the concept of shadow reward, we experiment with the single-agent setup. We consider the \textbf{exploration maximization} problem for the \emph{MountainCar} environment in which the two dimensional continuous state space is divided into $[12,11]$ grid size. We run the proposed algorithm for $40$ epochs and then plot the count based occupancy measure estimate in the first  row of Fig. \ref{fig1_entropy11}. In the figure, light color denotes lower value and dark color represent the higher values as shown in the colorbar.
We see that as we go from epoch $1$ to epoch $39$, the algorithm yields occupancy measures that better cover the state space, which is  achieved by the special structure of the ``shadow reward" we define as a by-product of the general utility.

%


\subsection{Multi-Agent Experiments}
For multi-agent problems, we experiment with $N\geq 2$ agents moving in a two-dimensional continuous space associated with the problem of \emph{Cooperative navigation} \cite{lowe2017multi}. 

\textbf{Exploration Maximization.} We consider a variant of the \emph{cooperative navigation} multi-agent environment provided in \cite{lowe2017multi} for $N=2$ agents. The goal of maximum entropy exploration in the multi-agent setting is one in which all agents in the network seek to cover the unknown space, whereby their local utility is the entropy in \eqref{defn:global-utility} is given by {$F_i(\lambda_{(i)}^\pi)\!=\!-\!\sum_{s{(i)}}\!\lambda_{(i)}^\pi(\!s_{(i)}\!)\!\cdot\!\log(\lambda_{(i)}^\pi(s_{(i)})\!)$.}

%
We compare DSAC against its corresponding centralized implementations (Cen-AC) or a variant that uses Monte-Carlo rollouts (Cen-R, Dec-R), as well as existing MaxEnt \cite{hazan2018provably} in Fig.\ref{fig_entropy_cost}-\ref{fig_entropy_occupancy}. See Appendix \ref{additional_multi} for details. Observe that MaxEnt does not achieve comparable performance, and DSAC achieves comparable performance to its variants that require centralization. Fig. \ref{fig_entropy_occupancy} visualizes the heatmap of the marginalized measure at agent 1 for DSAC (red) at different epochs as compared to MaxEnt (purple) and random baseline (green) -- note the superior space coverage of DSAC (red).

\begin{figure}[h]
	\vspace{-0mm}
	\centering
	\subfigure[ World model]{\includegraphics[width=0.25\columnwidth,height=0.14\textheight]{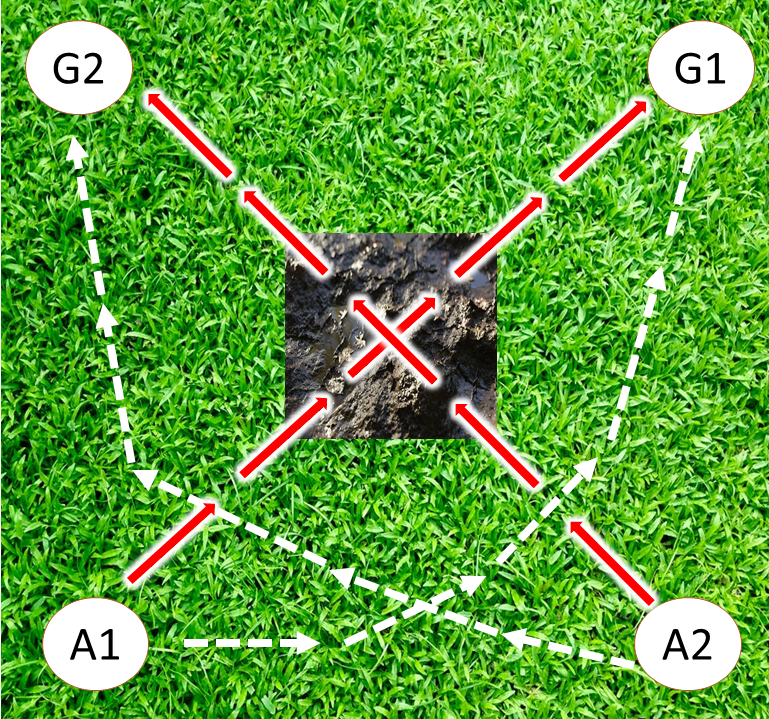}\label{fig_environment_multi}}
	\subfigure[ Average return]{\includegraphics[scale=0.31]{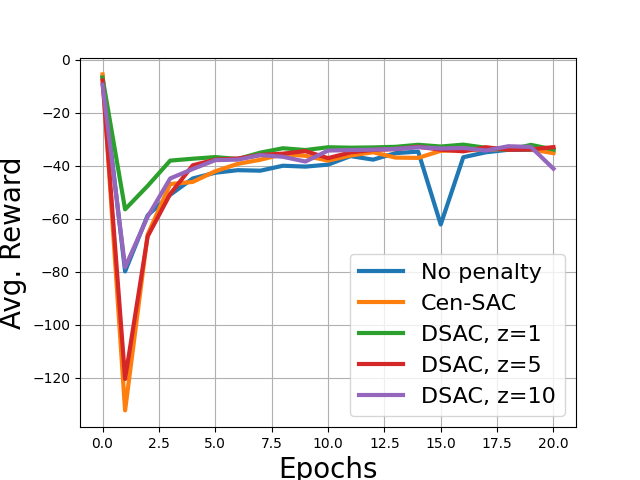}\label{fig_multi_reward}}
	\hspace{0cm}
	\subfigure[ Average cost]{\includegraphics[scale=0.31]{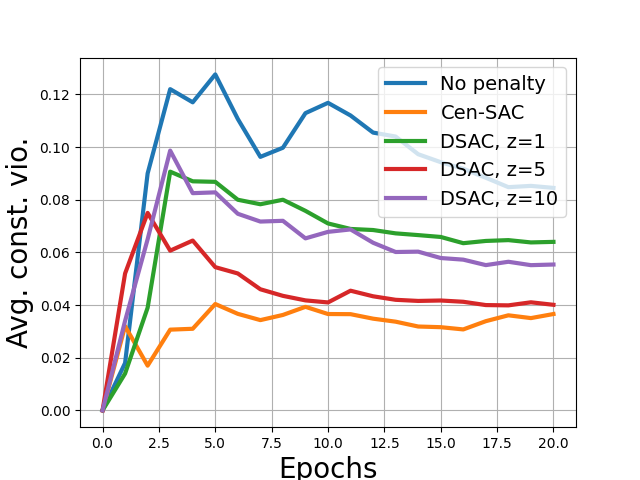}\label{fig_multi_constraint}}
	\vspace{-0mm}	\captionof{figure}{ (a) Two agent safe navigation environment with green as safe and brown as unsafe state space. The goal is to reach to goals (G1 and G2) safely from the starting positions (A1 and A2), respectively. (b) Undiscounted average reward return comparison and (c) average constraint violation comparison for different values of penalty parameter $z$. Observe that imposing constraints allows agents to avoid collision and the unsafe region, while effectively reaching their goals more often in terms of cumulative return and constraint violation.}
	\vspace{0mm}
\end{figure}

\begin{figure}[t!]
	\centering
	\subfigure[Average return]{\includegraphics[scale=0.30]{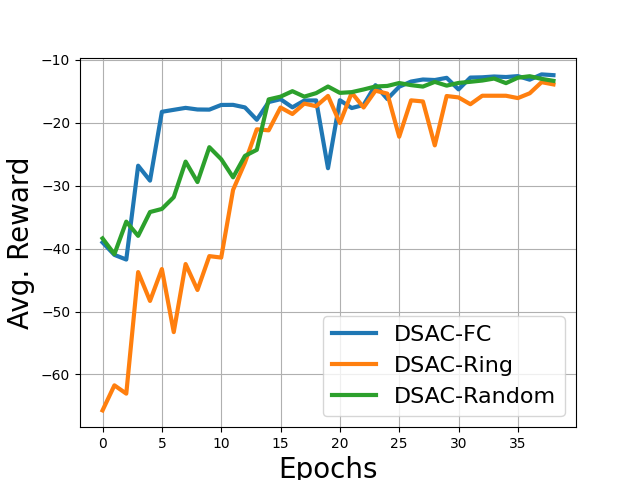}\label{fig_reward_LS}}\hspace{-5mm}
	\subfigure[Average cost]{\includegraphics[scale=0.30]{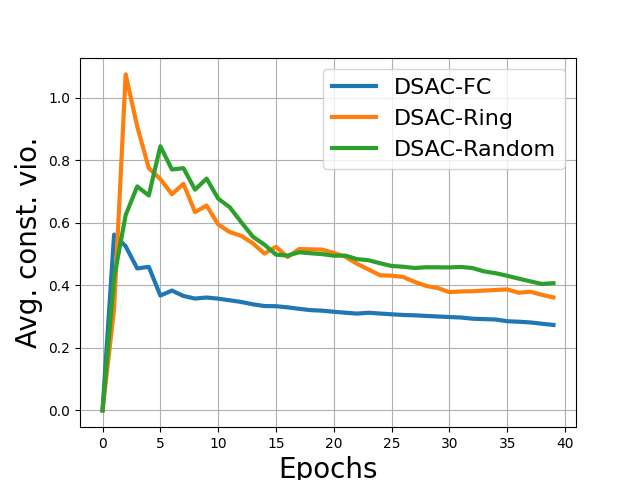}\label{fig_cost_LS}}\hspace{-5mm}
	\hspace{0cm}
	\subfigure[Consensus error]{\includegraphics[scale=0.30]{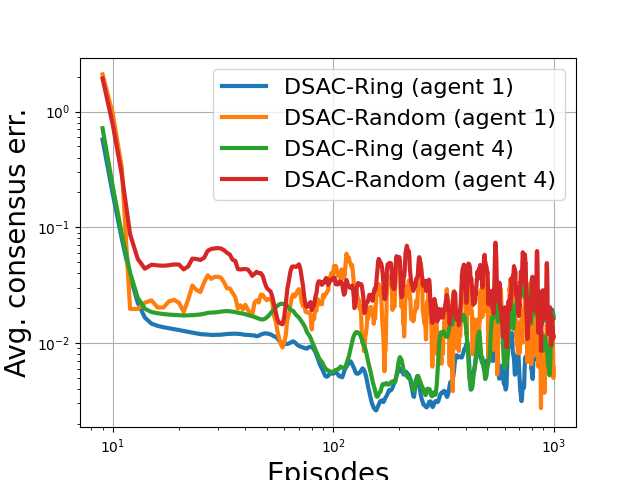}\label{fig_consensus}}\vspace{-0mm}
	\captionof{figure}{  {Safe navigation in a multi-agent cooperative environment with $4$ agents and $4$ landmarks. Note that the state space in this case would be $16$ dimensional (location of agent and landmarks). We run this experiment for three different communication graphs among agents;  \emph{fully connected (FC) (all the agents are connected to each other)}, \emph{ring} (all the agents are connected using ring topology), and \emph{random} (where agents are randomly using Erd\H{o}s-R\'enyi random graph model). (a) Running average of the reward return, (b) running average of the constraint violation, and (c) running average of the consensus error for agent $1$ and agent $4$ for \emph{ring} and \emph{random} network connectivity. Note that consensus error is zero for the fully connected network or a 2 agent network.} }
	\vspace{0mm}
	\label{fig_LS}
\end{figure}

\textbf{Safe Cooperative Navigation.} We consider a two agent cooperative environment from \cite{lowe2017multi} where each agent needs to reach its assigned goal while traversing only through the safe region as visualized in Fig.\ref{fig_environment_multi}. Agents receive a negative reward proportional to its distance from the landmark, and an additional negative reward of $-1$ if agents collide. Additionally, each agents receive a high cost of $c=1$  if it passes through the unsafe region (middle of the state space)  -- see Fig. \ref{fig_environment_multi}. We impose safety via the constraint for each agent $\ip{\lambda^\pi_i,c} \leq C$ where $\lambda^\pi_i$ in the marginalized occupancy measure, and including the constraint as a quadratic penalty in a manner similar to \eqref{eq:penalty} -- see Appendix \ref{additional_multi} for further details.
%
To solve this problem, we compare the performance of DSAC for various values of its penalty parameter $z$ to its centralized variant, and a version of multi-agent actor-critic that only ignores the cost. Results for the average reward and constraint violation, respectively, are given in Fig. \ref{fig_multi_reward}-\ref{fig_multi_constraint}. The decentralized DSAC achieves comparable performance to its centralized variant, and outperforms existing alternatives, yielding effective learned behaviors for navigation in team settings. Demonstrations for larger networks with different connectivities are in Figure \ref{fig_LS} and Appendix \ref{appendix_experiments}.  

\section{Conclusions}
We contributed a conceptual basis for defining agents' behavior in cooperative MARL beyond the cumulative return  via nonlinear functions of their occupancy measure. This motivates defining ``shadow rewards" and DSAC, whose critic employs shadow value functions and weighted averaging. Its consistency and sample complexity was rigorously established. Further, experiments illuminated the upsides of general utilities for teams. Future work includes improving communications and sample efficiencies, connections to meta-learning, and allowing information asymmetry.

\bibliography{bibliography}
\bibliographystyle{plain}
\appendix
\section*{Appendix}
\section{Further Context}\label{sec:related_works}
We expand upon different axes of comparison in order to clarify the technical setting under consideration. Notably, there has been a surge of interest in MARL in recent years, which has led to disparate possible technical settings: the definition of the MDP transition dynamics and associated correlation of agents' trajectories; the observability of the state variable itself, the availability of computational resources at a centralized location, and the protocol by which agents exchange information. We proceed to describe each of these facets in turn.

{\bf \noindent Transition Model.}  A key question when formulating MARL is the definition of the transition model, which determines the availability of local of global trajectory information to individuals. A standard operating hypothesis is that all agents have knowledge of the global state and action, and that the joint policy of the team factorizes into a product of individual marginal policies, which is referred to as \emph{joint action learners} (JAL) \cite{lee2020optimization}. When a common reward is perceived by all agents, this problem can be solved without agent coordination \cite{claus1998dynamics}, i.e., RL agents are effectively independent.

{\bf \noindent Observability.}  When the joint action of the team is unavailable to an individual, or other trajectory information is occluded, then one must contend with partial observability, i.e., solve a partially observed MDP (POMDP) \cite{krishnamurthy2016partially}. This is the case in both the setting of ``independent learners" (ILs) where agents only know their local action \cite{lee2020optimization}, or more broadly when they observe an insufficient statistic of the state \cite{mahajan2016decentralized}. Policy search in POMDPs is intractable in general, and even the policy evaluation (planning) problem requires solving a Bayesian inference problem in a Hidden Markov Model \cite{bernstein2002complexity}. To address HMMs in general requires complicated Monte Carlo schemes \cite{nayyar2013decentralized,zhang2019online} in order to propagate state information, and hence few convergence results exist for policy search in POMDPs.

{\bf \noindent Training Paradigm.} To contend with the intractability of POMDPs, a number of experimentally focused works \cite{foerster2016learning,leibo2017multi,foerster2017stabilising,rashid2018qmix} propose pooling all possibly incomplete information at a centralized location, and then feed it into a policy search routine. This paradigm, called centralized training decentralized execution (CTDE), pre-supposes the viability of an oracle agent for which centralization is safe and tractable, which in cases that agents geographically dispersed may not hold. 

{\bf \noindent Communication Model.}  For this reason, in this work, we instead focus on the case of \emph{decentralized training} of JAL, i.e., when full state-action information is available, but agents' rewards and policy parameters are held locally private. This setting has been studied extensively in recent years, giving rise to multi-agent extensions of TD \cite{lee2018stochastic,doan2019finite}, Q-learning \cite{kar2013cal}, value iteration \cite{wai2018multi,qu2019value}, and actor-critic \cite{zhang2018fully}. In these works, the manner in which agents may communicate is given by a graph: each agent is represented as a node and may communicate with another when there is a link between them. This graph is typically assumed to be a design parameter, as in this work. A separate but related body of works seek to estimate the communications architecture when agents' behavior is fixed \cite{eccles2019biases,ahilan2020correcting,bachrach2020negotiating}, or begin with locality-based correlation models to derive dependencies on agents' local utilities \cite{qu2020scalable,lin2020distributed}.

{\bf \noindent Information Mixing.} Under the hypothesis that the communications graph is given, the JAL setting with rewards and policy parameters held locally, then most MARL information exchange protocols may be placed within a burgeoning literature on multi-agent optimization. Predominately, in the aforementioned references, a  weighted averaging step is employed in order to diffuse information between agents across time while optimizing their local utility \cite{nedic2009distributed,chen2012diffusion}. This scheme originates in flocking \cite{tanner2007flocking} and gossip protocols \cite{boyd2006randomized,shah2009gossip}, and may be interpreted as an approximate enforcement strategy for equality constraints of the parameters held by distinct agents \cite{shi2015extra}. Alternatives based upon Lagrangian relaxation such as primal-dual \cite{koppel2015saddle}, alternating direction method of multipliers (ADMM) \cite{boyd2011distributed}, and dual reformulations \cite{terelius2011decentralized} can more sharply enforce consensus; however, we opt for a primal-only approach to enforcing consensus for simplicity and its compatibility with Perron-Frobenius theory \cite{chung1997spectral}, which provides appropriate conditions on the mixing matrices such that consensus is assured.

\begin{figure*}[ht]
	\centering
	\subfigure{\includegraphics[scale=0.3]{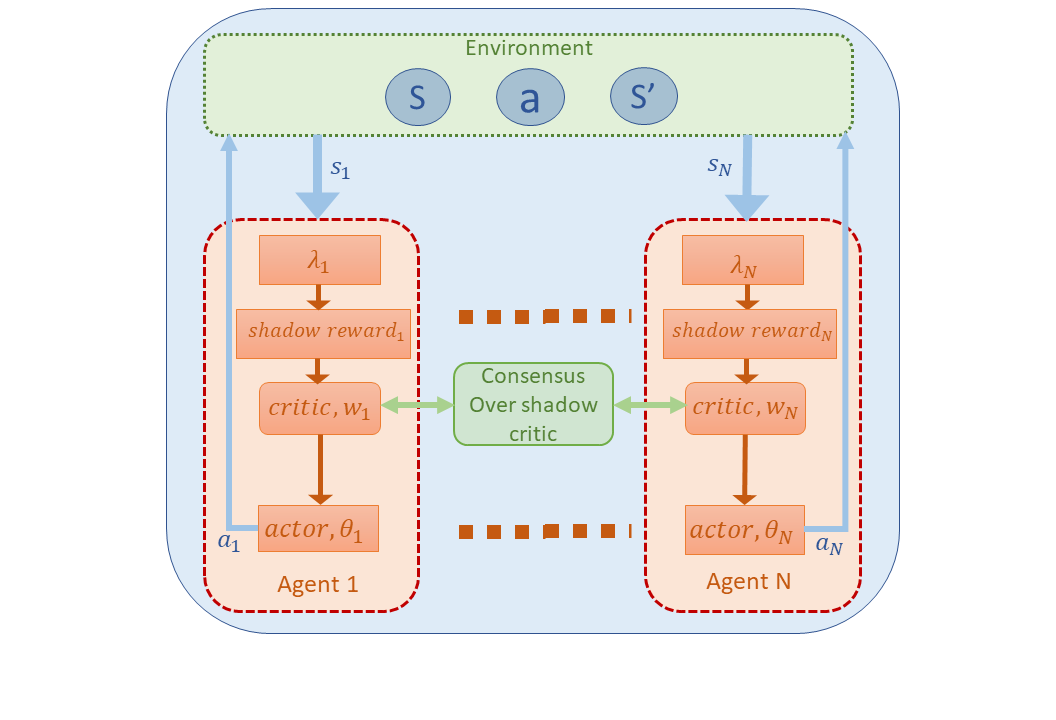}\label{fig1_entropy}}
	\captionof{figure}{The proposed DSCA algorthm execution.} 
	\label{algorithm_execution}
\end{figure*}
%
\section{Technical Details of Problem Setting}\label{apx:details}
\subsection{User Guide to Algorithm \ref{alg:stochastic-batch-2ts}}\label{apx:guide}
{A flow diagram explaining the algorithm execution is provided  in Fig. \ref{algorithm_execution}. Next we provide the detailed discussion of the proposed algorithm. }

\noindent \textbf{(i) Occupancy Measure Estimation.} With access to trajectory $\tau$, each agent $i$ seeks to evaluate its current policy in terms of the general utility. To do so, it must compute its shadow reward, a nonlinear function of the occupancy measure. This may be accomplished by first executing a local empirical occupancy measure estimator$ \hat\lambda^k_{i}$ by \eqref{eq:occupancy_measure_estimator}. And then one computes the shadow reward $\hat r^k_{i} = \nabla_{\lambda_{i}} F_{i}(\hat\lambda^k_{i})$.

\noindent \textbf{(ii) Shadow Policy Evaluation.} The shadow reward, as stated in \eqref{defn:reward-multi-agent-local}, is employed to formulate the local policy evaluation error $\ell_i$ [cf. \eqref{defn:critic-obj-marl-1}] with respect to the shadow $Q$ function \eqref{defn:Q-multi-agent-local}. Note that the shadow $Q$ function $Q^{\pi}_{F_i}(s^t,a^t)$ is substituted by an empirical estimate along the current trajectory. Specifically, $\hat Q^{t}_{i} = \sum_{t'=t}^{H}\gamma^{t'-t}\cdot r_i\big(s^{t'}_{(i)},a^{t'}_{(i)}\big)$ is the cumulative sum of tail rewards starting from $(s^t,a^t)$. Then, differentiating the resulting expression with respect to local critic parameters $w_i$ yields the critic gradient direction:
\begin{equation}
	\label{defn:Gw-1}
	\!\!\!\!\!G_{\!w_i}\!(\!\tau\!,\!r_i,\!w_i) \!=\! \!\sum_{t=0}^{H}\!\gamma^t\!\cdot\!(Q_{\!w_i}\!(\!s^{t},a^{t})\!-\!\hat Q^{t}_{i})\!\cdot\!\nabla_{\!w_i}Q_{w_i}\!(\!s^{t}\!,\!a^{t}),\!\!
\end{equation}
where agent $i$ then uses to update its local shadow critic as
$\hat w_{i}^{k+1} = w_{i}^k - \eta_w^k\cdot\widehat{\Delta}_{w_{i}}^k$
at step $k+1$, under the initialization with $w_1^0 = \cdots w_N^0$ and step-size $\eta_w^k$ specified as in Theorem \ref{theorem:final}. Moreover, $\widehat{\Delta}_{w_{i}}^k$ is a mini-batched version of the stochastic gradient in \eqref{defn:Gw-1} specified in Algorithm \ref{alg:stochastic-batch-2ts}.

\noindent \textbf{(iii) Information Exchange.} To ensure information effectively propagates across the network $\mathcal{G}$, agents perform a simple weighted averaging step using mixing matrix $M$, which is a symmetric doubly stochastic matrix that respects the edge connectivity of the graph, see Assumption \ref{assumption:mixing-matrix} for details. 
When agents execute $m$-steps of averaging per step $k$, we compactly express it as
$W^{k+1} \leftarrow M^m\cdot W^{k+1}.$
%


{\bf \noindent (iv) Policy Update.} Given the Q-function approximation parameter $w_i$, the actor gradient is constructed as
\begin{equation}
	\label{defn:Gtheta-1}
	G_{\theta_i}(\tau,w_i) \!=\! \sum_{t=0}^{H}\!\!\gamma^t Q_{w_i}(s^{t},a^{t})\nabla_{\theta_i} \log \pi_{\theta_i}^{(i)}(a^{t}_{(i)}|s^{t}).
\end{equation}
which is a stochastic approximation of the gradient in \eqref{eq:policy_gradient}.
Notice that replacing $Q_{w_i}(s^{t},a^{t})$ with the exact shadow Q-function $Q_F(s^{t},a^{t})$ reduces \eqref{defn:Gtheta-1}  to the REINFORCE \cite{williams1992simple} estimator equiped with the newly defined shadow Q-function. Then, each agent executes a simple mini-batch stochastic gradient ascent step.
\section{Proof of Lemma \ref{lemma:sufficient-ascent}}\label{appendix_taylor}
\begin{proof}
	By the $L_\theta$-smoothness of $F(\lambda^{\pi_{\theta}})$ to write
	\begin{align*}
		F(\lambda^{\pi_{\theta^{k+1}}}) &\geq F(\lambda^{\pi_{\theta^{k}}}) + \eta_\theta^k\sum_{i=1}^N\big\langle \nabla_{{\theta_i}} F(\lambda^{\pi_{\theta^k}}), \widehat{\Delta}_{\theta_{i}}^k\big\rangle - \frac{L_\theta(\eta_\theta^k)^2}{2}\sum_{i=1}^N\|\widehat{\Delta}_{\theta_{i}}^k\|^2\\
		& = F(\lambda^{\pi_{\theta^{k}}}) + \eta_\theta^k\sum_{i=1}^N\big\langle \nabla_{{\theta_i}} F(\lambda^{\pi_{\theta^k}}), \nabla_{{\theta_i}} F(\lambda^{\pi_{\theta^k}}) - \nabla_{{\theta_i}} F(\lambda^{\pi_{\theta^k}}) +  \widehat{\Delta}_{\theta_{i}}^k\big\rangle \\
		&  \quad- \frac{L_\theta(\eta_\theta^k)^2}{2}\sum_{i=1}^N\|\nabla_{{\theta_i}} F(\lambda^{\pi_{\theta^k}}) - \nabla_{{\theta_i}} F(\lambda^{\pi_{\theta^k}}) + \widehat{\Delta}_{\theta_{i}}^k\|^2\\
		& \geq F(\lambda^{\pi_{\theta^{k}}}) \!+\! \left(\frac{\eta_\theta^k}{2} \!-\! L_\theta(\eta_\theta^k)^2\right)\!\left\|\nabla_{\theta}F(\lambda^{\pi_{\theta^{k}}})\right\|^2 \!-\! \left(\frac{\eta_\theta^k}{2} \!+\! L_\theta(\eta_\theta^k)^2\right)\!\!\sum_{i=1}^{N}\left\|\nabla_{{\theta_i}}F(\lambda^{\pi_{\theta^{k}}})\!-\!\widehat{\Delta}_{\theta_i}^k\right\|^2.
	\end{align*}
	Then, by specifying the actor step-size as $\eta_\theta^k\leq 1/4L_\theta$, we obtain the statement in Lemma \ref{lemma:sufficient-ascent}.
\end{proof}

\section{Proof of Lemma \ref{lemma:stochastic-grad-err}}
\label{appdx:sto-grad-err}
\begin{proof}
	For simplicity, we denote $\cF_{k-1}$ as the $\sigma$-algebra generated by all the trajectories $\cup_{k'=0}^{k'=k-1}\cB_{k'}$. Before presenting the proof, let us provide the following lemma.
	\begin{lemma}
		\label{lemma:sto-grad-err-lm}
		For each agent $i$, in each iteration $k$, the estimator \eqref{eq:occupancy_measure_estimator} of the local occupancy measure [cf. \eqref{prop:local-measure-1}] associated with policy $\pi_{\theta^k}$ satisfies 
		\begin{eqnarray}
			\label{lm:stochastic-grad-err-1}
			\mathbf{Prob}\left(\|\hat\lambda^k_{i} - \lambda^{\pi_{\theta^{k}}}_{(i)}\|^2 \geq \frac{2\gamma^{2H_k}}{(1-\gamma)^2} + \frac{4+16\log\delta_k^{-1}}{(1-\gamma)^2B_k} \right)\leq \delta_k
		\end{eqnarray}
		As a result, with the shadow reward $r_i^k = \nabla_{\lambda_{(i)}} F\big(\lambda^{\pi_{\theta^k}}_{(i)}\big)$ as in \eqref{defn:reward-multi-agent-local}, we may write
		\begin{eqnarray}
			\label{lm:stochastic-grad-err-2}
			\mathbf{Prob}\left(\|\hat r^k_i - r^k_i\|^2_\infty \geq \frac{2L_\lambda^2\gamma^{2H_k}}{(1-\gamma)^2} + \frac{(4+16\log\delta_k^{-1})L_\lambda^2}{(1-\gamma)^2B_k} \right)\leq \delta_k
		\end{eqnarray}
	\end{lemma} 
	We move the proof of this lemma at the end of Appendix \ref{appdx:sto-grad-err}. 
	
	\textbf{1. Proof of Lemma \ref{lemma:stochastic-grad-err}\ref{lemma:stochastic-grad-err_1}}.
	First, for any $\tau\in\cB_k$, by direct computation, the estimation error of the actor update direction may be written as
	\begin{eqnarray*}
		\left\|G_{w_i}(\tau,\hat r^k_i,w_i^k) - G_{w_i}(\tau,r^k_i,w_i^k)\right\|^2
		&=&  \left\|\sum_{t=0}^{H}\gamma^t\cdot\sum_{t'=t}^{H}\gamma^{t'-t}(\hat r^k_i(s^{t},a^{t}) -  r^k_i(s^{t},a^{t}))\phi(s^{t},a^{t})\right\|^2\\
		&\leq&  \frac{C_\phi^2\|\hat r^k_i-r^k_i\|^2_\infty}{(1-\gamma)^4}.
	\end{eqnarray*}
	where we have computed the maximum over $(s,a)$ on the right-hand side, Cauchy-Schwartz, applied the boundedness of the feature representation (Assumption \ref{assumption:critc-features}), and applied the identity for a geometric sum defined by the discount factor. Let us denote $g_{w_i}^k = \frac{1}{B}\sum_{\tau\in\cB_k}G_{w_i}(\tau,r^k_i,w_i^k)$, then by the triangle inequality and the boundedness of the feature representation, we have
	we also have 
	\begin{eqnarray} 
		\label{lm:stochastic-grad-err-pf-2}
		\left\|\widehat{\Delta}_i^k- g_{w_i}^k\right\|^2& = & \left\|\frac{1}{B_k}\sum_{\tau\in\cB_k}G_{w_i}(\tau,\hat r^k_i,w_i^k)-\frac{1}{B_k}\sum_{\tau\in\cB_k}G_{w_i}(\tau,r^k_i,w_i^k)\right\|^2\nonumber\\
		&\leq &  \left(\frac{1}{B_k}\sum_{\tau\in\cB_k}\left\|G_{w_i}(\tau^{k,j},r^k_i,w_i^k)-G_{w_i}(\tau^{k,j},(r^k_i)^*,w_i^k)\right\|\right)^2\\
		& \leq & \left\|G_{w_i}(\tau,\hat r^k_i,w_i^k)-G_{w_i}(\tau,r^k_i,w_i^k)\right\|^2\nonumber\\
		& \leq & \frac{C_\phi^2\|\hat r^k_i-r^k_i\|^2_\infty}{(1-\gamma)^4},\nonumber
	\end{eqnarray} 
	with probability 1.  On the other hand, the magnitude of the directional error associated with the shadow critic  estimate may be written as
	\begin{align} 
		\|\nabla_{w_i} \ell_i(w_i^k;\pi_{\theta^k}) -& \EE\left[g_{w_i}^k|\cF_{k-1}\right]\|\nonumber
		\\
		= & \Bigg\|\EE\left[\sum_{t=0}^{+\infty}\gamma^t\cdot\left(\langle \phi(s^t,a^t),w_i^k\rangle-\sum_{t'=t}^{+\infty}\gamma^{t'-t}r_i^k(s^{t'},a^{t'})\right)\phi(s^t,a^t)\bigg| \xi,\pi_{\theta^k}\right] \nonumber
		\\
		& - \EE\left[\sum_{t=0}^{H_k}\gamma^t\cdot\left(\langle \phi(s^t,a^t),w_i^k\rangle-\sum_{t'=t}^{H}\gamma^{t'-t}r_i^k(s^{t'},a^{t'})\right)\phi(s^t,a^t)\bigg| \xi,\pi_{\theta^k}\right]\Bigg\|\nonumber\\
		\leq&   \Bigg\|\EE\left[\sum_{t=H_k+1}^{+\infty}\gamma^t\cdot\left(\langle \phi(s^t,a^t),w_i^k\rangle-\sum_{t'=t}^{+\infty}\gamma^{t'-t}r_i^k(s^{t'},a^{t'})\right)\phi(s^t,a^t)\bigg| \xi,\pi_{\theta^k}\right] \Bigg\|\nonumber
		\\
		& + \Bigg\|\EE\left[\sum_{t=0}^{H_k}\gamma^t\sum_{t'=H_k+1}^{+\infty}\gamma^{t'-t}r_i^k(s^{t'},a^{t'})\phi(s^t,a^t)\bigg| \xi,\pi_{\theta^k}\right]\Bigg\|\\
		\leq & \frac{\gamma^{H_k}\cdot C_\phi^2\|w_i^k\|}{1-\gamma} + \frac{\gamma^{H_k}\cdot C_\phi\|r_i^k\|_\infty}{(1-\gamma)^2} + \frac{\gamma^{H_k}\cdot H_kC_\phi}{1-\gamma}.
	\end{align} 
	where the first equality follows from adding and subtracting $\sum_{t=H_k+1}^{+\infty}[\gamma^t\sum_{t'=H_k+1}^{+\infty}\gamma^{t'-t}r_i^k(s^{t'},a^{t'}) ]\phi(s^t,a^t) $ inside the expectation, and the second applies the triangle inequality. The last follows from computing the maximum on the right-hand side, and applying the boundedness of the feature representation (Assumption \ref{assumption:critc-features}), and the identity for a geometric sum defined by the discount factor.
	Consequently, with probability 1, we have 
	\begin{eqnarray}
		\label{lm:stochastic-grad-err-pf-3}
		\left\|\nabla_{w_i} \ell_i(w_i^k;\pi_{\theta^k}) - \EE\left[g_{w_i}^k|\cF_{k-1}\right]\right\|^2\leq \frac{3C_\phi^2\gamma^{2H_k}}{(1-\gamma)^2}\cdot\left(C_\phi^2\|w_i^k\|^2  + H_k^2 + \frac{C_F^2}{(1-\gamma)^2}\right).
	\end{eqnarray}
	Now let us use concentration bound to control $\left\|g_{w_i}^k - \EE\left[g_{w_i}^k|\cF_{k-1}\right]\right\|^2$. Note that 
	\begin{eqnarray*}
		\|G_{w_i}(\tau,r^k_i,w_i^k)\|^2&=&\left\|\sum_{t=0}^{H_k}\!\gamma^t\cdot(Q_{\!w_i^k}\!(\!s^{t},a^{t})-\hat Q^{t}_{i})\cdot\nabla_{\!w_i}Q_{w_i^k}(s^{t},a^{t})\right\|^2\\
		& = & \left\|\sum_{t=0}^{H_k}\gamma^t\cdot(\phi(s^{t},a^{t})^\top(w_i^k)-\hat Q^{t}_{i})\cdot\phi(s^{t},a^{t})\right\|^2\\
		& \leq & C_\phi^2\left(\sum_{t=0}^{H_k}\gamma^t\cdot\Big(C_\phi\|w_i^k\|+\frac{C_F}{1-\gamma}\Big)\right)^2\\
		& \leq & \frac{2C_\phi^2}{(1-\gamma)^2}\cdot\left(C_\phi^2\|w_i^k\|^2 + \frac{C_F^2}{(1-\gamma)^2}\right).\\
	\end{eqnarray*} 
	Consequently, by Lemma 18 of \cite{kohler2017sub} and the definition of $g_{w_i}^k$, we have
	\begin{align}
		\label{lm:stochastic-grad-err-pf-4}
		\mathbf{Prob}\!\left(\!\left\|g_{w_i}^k \!-\! \EE\left[g_{w_i}^k|\cF_{k-1}\right]\right\|^2\geq \frac{(4+16\log(1/\delta_k))C_\phi^2}{B_k(1-\gamma)^2}\left(C_\phi^2\|w_i^k\|^2 + \frac{C_F^2}{(1-\gamma)^2}\!\right)\!\!\right)\! \leq \!\delta_k.
	\end{align}
	Now, the directional error associated with the critic with respect to $\widehat{\Delta}_{w_i}^k$ defined in Algorithm \ref{alg:stochastic-batch-2ts} together with basic inequality $(a+b+c)^2\leq 3(a^2+b^2+c^2)$ allows us to write:  
	\begin{align*}
		&\left\|\widehat{\Delta}_{w_i}^k - \nabla_{w_i}\ell_i(w_i^k;\pi_{\theta^k})\right\|^2 
		\nonumber
		\\
		&\quad= \left\|\widehat{\Delta}_{w_i}^k - g_{w_i}^k + g_{w_i}^k - \EE\left[g_{w_i}^k|\cF_{k-1}\right] + \EE\left[g_{w_i}^k|\cF_{k-1}\right]-\nabla_{w_i}\ell_i(w_i^k;\pi_{\theta^k})\right\|^2\\
		& \quad\leq 3\left\|\widehat{\Delta}_{w_i}^k - g_{w_i}^k\right\|^2+3\left\|g_{w_i}^k - \EE\left[g_{w_i}^k|\cF_{k-1}\right]\right\|^2 +3\left\|\EE\left[g_{w_i}^k|\cF_{k-1}\right]-\nabla_{w_i}\ell_i(w_i^k;\pi_{\theta^k})\right\|^2.
	\end{align*}    
	Combining \eqref{lm:stochastic-grad-err-pf-2}, \eqref{lm:stochastic-grad-err-pf-3}, \eqref{lm:stochastic-grad-err-pf-4}, \eqref{lm:stochastic-grad-err-2} and summing the inequality over all agents $i$ yields \eqref{lm:stochastic-grad-err-3}. \hfill \qed
	\\	
	\textbf{2. Proof of Lemma \ref{lemma:stochastic-grad-err}\ref{lemma:stochastic-grad-err_2}.}
	The proof of this inequality is similar to that of \eqref{lm:stochastic-grad-err-3}. First, 
	define the mini-batch gradient evaluated at the optimal critic parameters in the sense of \eqref{defn:critic-obj-marl-1} as $g_{\theta_i}^k = \frac{1}{B_k}\sum_{\tau\in\cB_k}G_{\theta_{i}}(\tau,w_{*}^{k+1})$. Then 
	\begin{align}
		\label{lm:stochastic-grad-err-pf-5}
		&\left\|\widehat{\Delta}_{\theta_i}^k - \nabla_{\theta_i} F(\lambda^{\pi_{\theta^{k}}})\right\|^2  =  \left\|\widehat{\Delta}_{\theta_i}^k \!-\! g_{\theta_i}^k \!+\! g_{\theta_i}^k \!-\! \EE[g_{\theta_i}^k|\cF_{k-1}] \!+\! \EE[g_{\theta_i}^k|\cF_{k-1}] \!-\! \nabla_{\theta_i} F(\lambda^{\pi_{\theta^{k}}})\right\|^2\\
		&\qquad\qquad\qquad\quad\,\,\,\,\qquad\leq  3\left\|\widehat{\Delta}_{\theta_i}^k \!-\! g_{\theta_i}^k\right\|^3 \!+\! 3\left\|g_{\theta_i}^k \!-\! \EE[g_{\theta_i}^k|\cF_{k-1}]\right\|^2 \!+\! 3\left\|\EE[g_{\theta_i}^k|\cF_{k-1}] \!-\! \nabla_{\theta_i} F(\lambda^{\pi_{\theta^{k}}})\right\|^2\nonumber.
	\end{align}
	where we add and subtract $ g_{\theta_i}^k$ and $\EE[g_{\theta_i}^k|\cF_{k-1}] $ and the basic inequality $(a+b+c)^2\leq 3(a^2+b^2+c^2)$.
	Now we bound the three terms one by one. For the first term, we substitute in the definition of $\widehat{\Delta}_{\theta_i}^k$ in Algorithm \ref{alg:stochastic-batch-2ts} and apply the triangle inequality, and upper-bound this expression by its maximum, making use of Proposition \ref{prop:lipschitz} and the boundedness of the score function (Assumption \ref{assumption:parameterization}) :
	\begin{eqnarray}
		\label{lm:stochastic-grad-err-pf-6}
		\left\|\widehat{\Delta}_{\theta_i}^k - g_{\theta_i}^k\right\|^2 & = & \left\|\frac{1}{B_k}\sum_{\tau\in\cB_k}G_{\theta_{i}}(\tau,w_{i}^{k+1})-\frac{1}{B_k}\sum_{\tau\in\cB_k}G_{\theta_{i}}(\tau,w_{*}^{k+1})\right\|^2\\
		& \leq & \frac{1}{B_k}\sum_{\tau\in\cB_k}\left\|G_{\theta_{i}}(\tau,w_{i}^{k+1})-G_{\theta_{i}}(\tau,w_{*}^{k+1})\right\|^2\nonumber\\
		& = & \left\|\sum_{t=0}^{H_k}\gamma^t\cdot\langle \phi(s^{t},a^{t}),w_i^{k+1}-w_*^{k+1}\rangle\cdot\nabla_{\theta_i} \log \pi_{\theta^k_i}^{(i)}(a^{t}_{(i)}|s^{t})\right\|^2\nonumber\\
		&\leq & \frac{C_\phi^2C_\pi^2}{(1-\gamma)^2}\cdot\|w_i^{k+1}-w_*^{k+1}\|^2\nonumber.
	\end{eqnarray}
	For the second term on the right-hand side of \eqref{lm:stochastic-grad-err-pf-5}, by Proposition \ref{prop:lipschitz} and the boundedness of the score function (Assumption \ref{assumption:parameterization}), we may write
	\begin{eqnarray*}
		\EE\left[\left\|G_{\theta_{i}}(\tau,w_{*}^{k+1})\right\|^2|\cF_{k-1}\right]
		& \!\!=\!\! & \EE\left[\left\|\sum_{t=0}^{H}\gamma^t\!\cdot\!\langle \phi(s^{t},a^{t}),w_*^{k+1}\rangle\nabla_{\theta_i} \log \pi_{\theta^k_i}^{(i)}(a^{t}_{(i)}|s^{t})\right\|^2\!\!|\cF_{k-1}\right]\\
		& \leq & \frac{C_\phi^2C_\pi^2\|w_*^{k+1}\|^2}{(1-\gamma)^2}.
	\end{eqnarray*}
	Then by Lemma 18 of \cite{kohler2017sub}, again, we have 
	\begin{eqnarray}
		\label{lm:stochastic-grad-err-pf-7}
		\mathbf{Prob}\left(\left\|g_{\theta_i}^k \!-\! \EE[g_{\theta_i}^k|\cF_{k-1}]\right\|^2\geq\frac{2(1+4\log(1/\delta_k))C_\phi^2C_\pi^2\|w_*^{k+1}\|^2}{(1-\gamma)^2B_k}\right)\leq \delta_k.
	\end{eqnarray}
	Finally, for the last term on the right-hand side of \eqref{lm:stochastic-grad-err-pf-5}, we have 
	\begin{eqnarray}
		\label{lm:stochastic-grad-err-pf-8}
		& &\sum_{i=1}^N\left\|\EE[g_{\theta_i}^k|\cF_{k-1}] \!-\! \nabla_{\theta_i} F(\lambda^{\pi_{\theta^{k}}})\right\|^2\\
		& = & \sum_{i=1}^N\left\|\EE\left[\sum_{t=0}^{H_k}\gamma^t\langle \phi(s^{t},a^{t}),w_*^{k+1}\rangle\cdot\nabla_{\theta_i} \log \pi_{\theta^k_i}^{(i)}(a^{t}_{(i)}|s^{t})\Big|\xi, \pi_{\theta^{k}}\right] \!-\! \nabla_{\theta_i} F(\lambda^{\pi_{\theta^{k}}})\right\|^2\nonumber\\
		& \leq & 2\sum_{i=1}^N\left\|\EE\left[\sum_{t=H_k+1}^{+\infty}\gamma^t\langle \phi(s^{t},a^{t}),w_*^{k+1}\rangle\cdot\nabla_{\theta_i} \log \pi_{\theta^k_i}^{(i)}(a^{t}_{(i)}|s^{t})\Big|\xi, \pi_{\theta^{k}}\right]\right\|^2\nonumber\\
		& & + 2\underbrace{\sum_{i=1}^N\left\|\EE\left[\sum_{t=0}^{+\infty}\gamma^t\langle \phi(s^{t},a^{t}),w_*^{k+1}\rangle\cdot\nabla_{\theta_i} \log \pi_{\theta^k_i}^{(i)}(a^{t}_{(i)}|s^{t})\Big|\xi, \pi_{\theta^{k}}\right] \!-\! \nabla_{\theta_i} F(\lambda^{\pi_{\theta^{k}}})\right\|^2}_{E_{\theta^{k}}^2}\nonumber\\
		& \leq & \frac{2\gamma^{2H_k}C_\phi^2C_\pi^2N\|w_*^{k+1}\|^2}{(1-\gamma)^2} + 2E_{\theta^{k}}^2\nonumber.
	\end{eqnarray}
	where we have used the triangle inequality, Proposition \ref{prop:lipschitz}, the boundedness of the score function (Assumption \ref{assumption:parameterization}), and the boundedness of the feature representation (Assumption \ref{assumption:critc-features}).
	Combining \eqref{lm:stochastic-grad-err-pf-5}, \eqref{lm:stochastic-grad-err-pf-6}, \eqref{lm:stochastic-grad-err-pf-7} and \eqref{lm:stochastic-grad-err-pf-8} and summing over $i=1,...,N$ allows us to conclude the result of Lemma \ref{lemma:stochastic-grad-err}\ref{lemma:stochastic-grad-err_2}, with the random variable defined as 
	\begin{equation}\label{eq:zeta_theta}
		\zeta_\theta^k = 3\sum_{i=1}^N\left\|g_{\theta_i}^k \!-\! \EE[g_{\theta_i}^k|\cF_{k-1}]\right\|^2.
	\end{equation}
\end{proof}
\subsection{Proof of Lemma \ref{lemma:sto-grad-err-lm}}
\begin{proof}
	First, for the ease of notation, we denote the empirical occupancy measure estimate of agent $i$ along trajectory $\tau$ as $ \lambda_{i}(\tau):=\sum_{t=0}^H\gamma^t\cdot\be(s^t_{(i)},a^t_{(i)})$. Then it is easy to see that for any $\tau\in\cB_k$, one has
	\begin{align}
		\left\|\EE\left[\hat\lambda^k_{i}|\cF_{k-1}\right] -\lambda^{\pi_{\theta^k}}_{(i)}\right\|_1 =& \left\|\EE\left[{\lambda}_{i}(\tau)|\cF_{k-1}\right] -\lambda^{\pi_{\theta^k}}_{(i)}\right\|_1 \nonumber
		\\
		\leq & \frac{\gamma^{H_k}}{1-\gamma}.
	\end{align}
	Second, note that $\|{\lambda}_{i}(\tau)\|^2\leq \frac{1}{(1-\gamma)^2}$ almost surely. Then, by Lemma 18 of \cite{kohler2017sub}, we have
	\begin{eqnarray*}
		\mathbf{Prob}\left(\left\|\frac{1}{B_k}\sum_{\tau\in\cB_k}{\lambda}_{i}(\tau) - \EE\left[{\lambda}_{i}(\tau)\big|\cF_{k-1}\right]\right\|^2\geq\epsilon\right)\leq \exp\left\{-\frac{(1-\gamma)^2\epsilon \cdot B_k - 2}{8}\right\}
	\end{eqnarray*}
	By setting $\epsilon = \frac{2+8\log\delta^{-1}_k}{(1-\gamma)^2B_k}$, we have 
	\begin{eqnarray*}
		\mathbf{Prob}\left(\left\|\hat\lambda^k_{i} - \EE\left[{\lambda}_{i}(\tau)\big|\cF_k\right]\right\|^2\geq\frac{2+8\log\delta_k^{-1}}{(1-\gamma)^2B_k}\right)\leq \delta_k.
	\end{eqnarray*}
	Consequently, we have 
	\begin{align}
		\label{lm:stochastic-grad-err-pf-1}
		\left\|\hat\lambda^k_{i}-\lambda^{\pi_{\theta^{k}}}_{(i)}\right\|^2 \leq \frac{2\gamma^{2H_k}}{(1-\gamma)^2} + \frac{4+16\log\delta_k^{-1}}{(1-\gamma)^2B_k} \quad\mbox{w.p.} \quad 1-\delta_k,
	\end{align} 
	which completes the proof. 
\end{proof}

\section{Proof of Lemma \ref{lemma:marl-dis-from-avg-sto}}
\label{appdx:lemma-marl-dis-from-avg-sto}
\begin{proof}
	First, for the mixing matrix $M$, we introduce the following supporting lemma, whose proof is straightforward and is hence omitted.  
	\begin{lemma}
		\label{lemma:mixing-matrix}
		Let the mixing matrix $M$ satisfy the Assumption \ref{assumption:mixing-matrix}, then  
		$$\left\|M^k - \frac{1}{N}\cdot\mathbf{1}_N\mathbf{1}_N^\top\right\|_2\leq \rho^k, \forall k\geq1$$ 
		where $\|\cdot\|_2$ stands for the spectral norm of a matrix.
	\end{lemma}
	This result is standard in multi-agent optimization and spectral graph theory -- see \cite{nedic2009distributed}[Proposition 1 and 2].
	For the ease of notation, let us work with the following matrix version of the critic objective function as well as the update of the Algorithm \ref{alg:stochastic-batch-2ts}. Denote $\Phi\in\RR^{d\times |\cS||\cA|}$ be the feature matrix, with the $(s,a)$-th column being $\phi(s,a)$. Denote $\Lambda_k:=\mathrm{Diag}(\lambda^{\pi_{\theta^{k}}})$ as a diagonal matrix in $\RR^{|\cS||\cA|\times|\cS||\cA|}$. Denote $Q^{\pi_{\theta^{k}}}_F\in\RR^{|\cS||\cA|}$ as the vectorization of the global shadow Q-function $Q^{\pi_{\theta^{k}}}_F(\cdot)$. Similarly, we also denote $Q^{\pi_{\theta^{k}}}_{i}\in\RR^{|\cS||\cA|}$ to be the vectorization of the local shadow Q-functions $Q^{\pi_{\theta^{k}}}_{i}(\cdot)$ defined in \eqref{defn:Q-multi-agent-local}. Finally, by stacking the vectors $Q^{\pi_{\theta^{k}}}_{i}$ together, we denote the matrix $\mathbf{Q}^{\pi_{\theta^{k}}} = \left[Q^{\pi_{\theta^{k}}}_{1},Q^{\pi_{\theta^{k}}}_{2},\cdots,Q^{\pi_{\theta^{k}}}_{N}\right]\in\RR^{|\cS||\cA|\times N}$. Then by definition, we have $Q^{\pi_{\theta^{k}}}_F = \mathbf{Q}^{\pi_{\theta^{k}}}\cdot\frac{\mathbf{1}_N}{N}$, where $\mathbf{1}_N$ is an N-dimensional all-one column vector. With those definitions, we can rewrite the critic objective and its gradient as 
	\begin{equation}
		\label{lm:marl-dis-from-avg-1}
		\ell(w;\pi_{\theta^{k}}) = \frac{1}{2}\left\|\Phi^\top w-Q^{\pi_{\theta^{k}}}_F\right\|^2_{\Lambda_k}\quad\mbox{and}\quad\nabla_w\ell(w;\pi_{\theta^{k}}) = \Phi\Lambda_k(\Phi^\top w-Q^{\pi_{\theta^{k}}}_F).
	\end{equation}
	\begin{equation}
		\label{lm:marl-dis-from-avg-2}
		\ell_i(w_i;\pi_{\theta^{k}}) = \frac{1}{2}\left\|\Phi^\top w_i-Q^{\pi_{\theta^{k}}}_{i}\right\|^2_{\Lambda_k}\quad\mbox{and}\quad\nabla_{w_i}\ell_i(w_i;\pi_{\theta^{k}}) = \Phi\Lambda_k(\Phi^\top w_i-Q^{\pi_{\theta^{k}}}_{i}).
	\end{equation}
	Therefore, the update of the critic variables can be written as 
	\begin{equation}
		\label{lm:marl-dis-from-avg-sto-3}
		W^{k+1} = \left(W^k - \eta_w^k\Phi\Lambda_k\left(\Phi^\top W^k - \mathbf{Q}^{\pi_{\theta^{k}}}\right) + \eta_w^k\zeta_W^k\right)\cdot M^m. 
	\end{equation}
	Multiplying both sides of the above equation with $\frac{1}{N}\cdot \mathbf{1}_N$ yields
	\begin{eqnarray*} 
		\bar w^{k+1} & = & \frac{1}{N}\cdot W^{k+1} \cdot\mathbf{1}_N \\
		& = & \left(W^k - \eta_w^k\Phi\Lambda_k\left(\Phi^\top W^k - \mathbf{Q}^{\pi_{\theta^{k}}}\right)+ \eta_w^k\zeta_W^k\right)\cdot \frac{M^m\cdot\mathbf{1}_N}{N}\\
		& = & \left(W^k - \eta_w^k\Phi\Lambda_k\left(\Phi^\top W^k - \mathbf{Q}^{\pi_{\theta^{k}}}\right)+ \eta_w^k\zeta_W^k\right)\cdot \frac{\mathbf{1}_N}{N}\\
		& = & \bar w^k - \eta_w^k\Phi\Lambda_k(\Phi^\top\bar w^k-Q^{\pi_{\theta^{k}}})+ \frac{\eta_w^k\zeta_W^k\mathbf{1}_N}{N}.
	\end{eqnarray*}
	That is, the average of the local updates satisfies:
	\begin{equation}
		\label{lm:marl-dis-from-avg-sto-4}
		\bar w^{k+1} = \bar w^k - \eta_w^k\Phi\Lambda_k(\Phi^\top\bar w^k-Q^{\pi_{\theta^{k}}}) + \frac{\eta_w^k\zeta_W^k\mathbf{1}_N}{N}.
	\end{equation}
	Consider the difference between $W^{k+1}$ and $\bar w^{k+1}$ respectively in \eqref{lm:marl-dis-from-avg-sto-3} and \eqref{lm:marl-dis-from-avg-sto-4}:
	\begin{align*}
		&\!\!\!\!\!W^{k+1} - \bar w^{k+1}\cdot \mathbf{1}_N^\top\\ & = \left(W^k - \eta_w^k\Phi\Lambda_k\left(\Phi^\top W^k \!-\! \mathbf{Q}^{\pi_{\theta^{k}}}\right)\!+\!\eta_w^k\zeta_W^k\right)\!\cdot\! M^m \!-\! \left(\bar w^k \!-\! \eta_w^k\Phi\Lambda_k(\Phi^\top\bar w^k\!-\!Q^{\pi_{\theta^{k}}})\!+ \frac{\eta_w^k\zeta_W^k\mathbf{1}_N}{N}\!\right)\mathbf{1}_N^\top\nonumber\\
		\!&= \left(W^k \!-\! \eta_w^k\Phi\Lambda_k\left(\Phi^\top W^k \!-\! \mathbf{Q}^{\pi_{\theta^{k}}}\right)\!+\!\eta_w^k\zeta_W^k\right)\!\cdot\! M^m - \left(\bar w^k \!-\! \eta_w^k\Phi\Lambda_k(\Phi^\top\bar w^k\!-\!Q^{\pi_{\theta^{k}}})\!+ \!\frac{\eta_w^k\zeta_W^k\mathbf{1}_N}{N}\!\right)\mathbf{1}_N^\top M^m\nonumber\\
		\!& =  \left(I\!-\!\eta_w^k\Phi\Lambda_k\Phi^\top\right)\left(W^k \!-\!\bar w^k\cdot\mathbf{1}_N^\top \right)\!\cdot\! M^m  \!+\! \eta_w^k\Phi\Lambda_k\!\left(Q^{\pi_{\theta^{k}}}\mathbf{1}_N^\top \!-\! \mathbf{Q}^{\pi_{\theta^{k}}}\right)\!M^m \! - \!\eta_w^k\zeta_W^k\!\left(\!\frac{\mathbf{1}_N\mathbf{1}_N\top}{N}\!-\!M^m\right)\\
		\!&= \left(I\!-\!\eta_w^k\Phi\Lambda_k\Phi^\top\!\right)\!\left(\!W^k \!	\!-\!\bar w^k\cdot\mathbf{1}_N^\top \right)\!\cdot\! M^m  \!+\! \eta_w^k\Phi\Lambda_k\left(\mathbf{Q}^{\pi_{\theta^{k}}}\!\cdot\!\frac{1}{N}\mathbf{1}_N\mathbf{1}_N^\top- \mathbf{Q}^{\pi_{\theta^{k}}}\right)\!M^m - \eta_w^k\zeta_W^k\!\left(\!\frac{\mathbf{1}_N\mathbf{1}_N\top}{N}\!-\!M^m\!\right)\\
		\!& =  \left(I-\eta_w^k\Phi\Lambda_k\Phi^\top\right)\left(W^k \!-\!\bar w^k\cdot\mathbf{1}_N^\top \right)\!\cdot\! M^m  + \eta_w^k(\Phi\Lambda_k\mathbf{Q}^{\pi_{\theta^{k}}}-\zeta_W^k)\left(\frac{1}{N}\mathbf{1}_N\mathbf{1}_N^\top - M^m\right).
	\end{align*}
	In the second equality, we exploit the fact that $\mathbf{1}_N$ is an eigenvector of $M$, in the third we group like terms, and in the fourth we again use the fact that $\mathbf{1}_N$ is an eigenvector of $M$, and lastly we again group like terms. Let us repeat the above recursion backwards  $k'=k, k-1, \dots,0$, and use the fact that $W^0-\bar w^0\mathbf{1}_N^\top = 0$ due to our initialization. Then we have 
	\begin{align}
		\label{lm:marl-dis-from-avg-sto-5}
		W^{k+1} - \bar w^{k+1}\mathbf{1}_N^\top =  \sum_{k'=0}^{k} \left(\Pi^{k}_{t=k'+1}(I\!-\!\eta_w^t\Phi\Lambda_t\Phi^\top)\!\right)\!\cdot\! \eta_w^{k'}(\Phi\Lambda_{k'}\mathbf{Q}^{\pi_{\theta^{k'}}}-\zeta_W^{k'})\!\left(\frac{1}{N}\mathbf{1}_N\mathbf{1}_N^\top - \!M^{m(k-k'+1)}\right).
	\end{align}
	For the above equality, because $\eta_w^t\leq 1/L_w$, we have for the first factor on the right-hand side of the preceding expression
	\begin{equation}
		\label{lm:marl-dis-from-avg-sto-6}
		\left\|\Pi^{k}_{t=k'+1}(I\!-\!\eta_w^t\Phi\Lambda_t\Phi^\top)\right\|_2\leq \Pi^{k}_{t=k'+1}(1\!-\!\eta_w^t\mu_w)\leq 1.
	\end{equation}
	On the other hand, the later two factors on the right-hand side of \eqref{lm:marl-dis-from-avg-sto-5} may be upper-estimated by the gradient estimation error $\zeta_W^k$ [defined in Lemma \ref{lemma:stochastic-grad-err}] and problem-dependent constants as
	\begin{eqnarray}
		\label{lm:marl-dis-from-avg-sto-7}
		& & \left\|\eta_w^{k'}(\Phi\Lambda_{k'}\mathbf{Q}^{\pi_{\theta^{k'}}}-\zeta_W^{k'})\!\left(\frac{1}{N}\mathbf{1}_N\mathbf{1}_N^\top - \!M^{m(k-k'+1)}\right)\right\|_F\nonumber\\
		& \leq & \eta_w^{k'}\left\|\left(\frac{1}{N}\mathbf{1}_N\mathbf{1}_N^\top - M^{m(k-k'+1)}\right)\right\|_2\cdot\left\|\Phi\Lambda_{k'}\mathbf{Q}^{\pi_{\theta^{k'}}}-\zeta_W^{k'}\right\|_F\nonumber\\
		& \leq & \eta_w^{k'}\left\|\left(\frac{1}{N}\mathbf{1}_N\mathbf{1}_N^\top - M^{m(k-k'+1)}\right)\right\|_2\cdot\left(\left\|\Phi\Lambda_{k'}\mathbf{Q}^{\pi_{\theta^{k'}}}\right\|_F + \|\zeta_W^{k'}\|_F\right)\\
		&\leq & \eta_w^{k'}\rho^{m(k-k'+1)}\left(\left\|\Phi\Lambda_{k'}\mathbf{Q}^{\pi_{\theta^{k'}}}\right\|_F + \|\zeta_W^{k'}\|_F\right)\nonumber\\
		& = & \eta_w^{k'}\rho^{m(k-k'+1)}\left(\|\zeta_W^{k'}\|_F + \sqrt{\sum_{i=1}^N\left\|\sum_{s,a} \lambda^{\pi_{\theta^{k'}}}(s,a)\cdot Q^{\pi_{\theta^{k'}}}_{i}(s,a)\cdot\phi(s,a) \right\|^2}\right)\nonumber\\
		&\overset{(a)}{\leq}& \eta_w^{k'}\rho^{m(k-k'+1)}\left(\|\zeta_W^k\|_F +\sqrt{\sum_{i=1}^N\left(\sum_{s,a} \lambda^{\pi_{\theta^{k'}}}(s,a)\right)^2\frac{C_F^2C_\phi^2}{(1-\gamma)^2}}\right)\nonumber\\
		& \leq & \left(\|\zeta_W^{k'}\|_F + \frac{\sqrt{N}C_FC_\phi}{(1-\gamma)^2}\right)\eta_w^{k'}\rho^{m(k-k'+1)}\nonumber
	\end{eqnarray}
	where the first inequality applies Cauchy-Schwartz, the second employs the triangle inequality. Moreover, the third inequality makes use of Lemma \ref{lemma:mixing-matrix} and the definition of $\zeta_W^k$ [see Lemma \ref{lemma:stochastic-grad-err}], and the fourth expands the definition of $ \left\|\Phi\Lambda_{k'}\mathbf{Q}^{\pi_{\theta^{k'}}}\right\|_F$. Additionally, (a) applies the fact that $Q_i^\pi\leq \frac{C_F}{1-\gamma}$, and the last inequality replaces the occupancy measure by its magnitude. Combining \eqref{lm:marl-dis-from-avg-sto-5}, \eqref{lm:marl-dis-from-avg-sto-6} and \eqref{lm:marl-dis-from-avg-sto-7} yields 
	\begin{eqnarray*} 
		\left\|W^{k+1} \!-\! \bar w^{k+1}\mathbf{1}_N^\top\right\| &\leq &  \left(\max_{k'\leq k}\|\zeta_W^{k'}\|_F + \frac{\sqrt{N}C_FC_\phi}{(1-\gamma)^2}\right)\rho^m\sum_{k'=0}^k\eta_w^{k'}\rho^{m(k-k')}.
	\end{eqnarray*}
	Squaring both sides of the above inequality yields the result of Lemma \ref{lemma:marl-dis-from-avg-sto}.
\end{proof} 

\section{Proof of Lemma \ref{lemma:marl-contraction-sto}}
\label{appdx:lm-marl-contraction-sto}
\begin{proof}
	First, for the ease of notation, we will use $w^{k+1}_*$ to denote $w^*(\theta^k)$.  The key observation that underlies our proof is the fact that \eqref{lm:marl-dis-from-avg-sto-4} and \eqref{lm:marl-dis-from-avg-1} taken in tandem may be reinterpreted as the averaged critic parameter performing gradient descent of the critic objective function $\ell(w;\pi_{\theta^{k}})$, with the error $\zeta_W^k$. The optimality condition of  $\ell(w;\pi_{\theta^{k}})$ may be rearranged to write
	$$\Phi\Lambda_kQ^{\pi_{\theta^{k}}} = \Phi\Lambda_k\Phi^\top w^{k+1}_*.$$ 
	Combining the above equation with \eqref{lm:marl-dis-from-avg-sto-4} yields
	$$\bar w^{k+1}-w^{k+1}_* = \left( I - \eta_w^k\Phi\Lambda_k\Phi^\top\right)(\bar w^k - w^{k+1}_*) + \frac{\eta_w^k\zeta_W^k\mathbf{1}_N}{N}.$$
	Computing the magnitude of both sides and applying \eqref{lm:marl-dis-from-avg-sto-6} to $( I - \eta_w^k\Phi\Lambda_k\Phi^\top)$ yields:
	\begin{equation}
		\label{lm:marl-contraction-sto-1}
		\|\bar w^{k+1}-w^{k+1}_*\| \leq \left (1-\eta_w^k\mu_w\right)\|\bar w^k - w^{k+1}_*\| + \frac{\eta_w^k\|\zeta_W^k\|_F}{\sqrt{N}}.
	\end{equation}
	Consequently, squaring both sides, grouping like terms, and adding and subtracting $ w^{k}_*$ inside the norm-difference term yields
	\begin{align}
		\label{lm:marl-contraction-sto-2}
		\|\bar w^{k+1}-w^{k+1}_*\|^2&= \left(1-\eta_w^k\mu_w\right)^2\left(1+\eta_w^k\mu_w\right)\|\bar w^k - w^{k+1}_*\|^2 + \left(1+\frac{1}{\eta_w^k\mu_w}\right)\frac{(\eta_w^k)^2\|\zeta_W^k\|_F^2}{N}\nonumber\\
		& \leq   \left(1-\eta_w^k\mu_w\right)\|\bar w^k - w^{k}_* + w^{k}_*-w^{k+1}_*\|^2+ \frac{2\eta_w^k\|\zeta_W^k\|_F^2}{\mu_w\cdot N}\nonumber\\
		& \leq  \left(1-\eta_w^k\mu_w\right)\left(1+\frac{\eta_w^k\mu_w}{2}\right)\|\bar w^k - w^{k}_* \|^2 \nonumber\\
		&\qquad+ \left(1-\eta_w^k\mu_w\right)\left(1+\frac{1}{2\eta_w^k\mu_w}\right)\|w^{k}_*-w^{k+1}_*\|^2+ \frac{2\eta_w^k\|\zeta_W^k\|_F^2}{\mu_w\cdot N}\nonumber\\
		&\leq  \left(1-\frac{\eta_w^k\mu_w}{2}\right)\|\bar w^k - w^{k}_* \|^2 + \frac{1}{\eta_w^k\mu_w}\|w^{k}_*-w^{k+1}_*\|^2+ \frac{2\eta_w^k\|\zeta_W^k\|_F^2}{\mu_w\cdot N}
	\end{align}
	where we frequently use the fact that $(a+b)^2\leq (1+c)a^2 + (1+c^{-1})b^2$ for any $c>0$. 
	Next, we analyze the second term on the right-hand size of \eqref{lm:marl-contraction-sto-2}.
	\begin{align}
		\label{lm:marl-contraction-sto-3}
		\|w^{k}_*-w^{k+1}_*\|^2
		&\leq  C_w^2\|\theta^{k}-\theta^{k-1}\|^2 \leq  C_w^2(\eta_\theta^{k-1})^2\sum_{i=1}^{N}\|\widehat{\Delta}_{\theta_i}^{k-1}\|^2\nonumber\\
		& \leq 2C_w^2(\eta_\theta^{k-1})^2\sum_{i=1}^{N}\|\nabla_{{\theta_i}} F(\lambda^{\pi_{\theta^{k-1}}})\|^2 + 2C_w^2(\eta_\theta^{k-1})^2\sum_{i=1}^{N}\|\nabla_{{\theta_i}} F(\lambda^{\pi_{\theta^{k-1}}}) - \widehat{\Delta}_{\theta_i}^{k-1}\|^2\nonumber\\
		& \leq  2C_w^2(\eta_\theta^{k-1})^2\Bigg(\!\|\nabla_{{\theta}} F(\lambda^{\pi_{\theta^{k-1}}}\!)\!\|^2  \nonumber\\
		&\qquad\hspace{2.2cm}+ \frac{3C_\phi^2C_\pi^2}{(1-\gamma)^2}\sum_{i=1}^{N}\|w_i^{k}\!-\!w_*^{k}\|^2+ \zeta_\theta^{k-1} +6E_{\theta^{k-1}}^2+\cO( \gamma^{2H_{k-1}})\!\Bigg)\nonumber\\
		&\leq   2C_w^2(\eta_\theta^{k-1})^2\Big(\|\nabla_{{\theta}} F(\lambda^{\pi_{\theta^{k-1}}})\|^2 + \zeta_\theta^{k-1} +6E_{\theta^{k-1}}^2+\cO( \gamma^{2H_{k-1}})  \nonumber \\
		&\qquad\qquad\qquad\qquad\quad+ \frac{6C_\phi^2C_\pi^2}{(1-\gamma)^2}\sum_{i=1}^{N}\left(\|\bar w^{k}-w^{k}_*\|^2+\| w^{k}_i-\bar w^{k}\|^2\right)\Big).
	\end{align}
	In the first line, we applied the smoothness of $w^{k+1}_*$ with respect to $\theta_k$ in defined in as a hypothesis of Lemma \ref{lemma:marl-contraction-sto}, the second inequality makes use of the definition of the update for $\theta^k$ in terms of the update direction $\widehat{\Delta}_{\theta_i}^{k-1}$ and recursively expands the sum followed by the triangle inequality. The third substitutes in the definition of the update $\widehat{\Delta}_{\theta_i}^{k-1}$ from Algorithm \ref{alg:stochastic-batch-2ts}, adds and subtracts $\nabla_{{\theta}} F(\lambda^{\pi_{\theta^{k-1}}})$ inside the norm and applies the triangle inequality. The fourth inequality applies Lemma \ref{lemma:stochastic-grad-err}\ref{lemma:stochastic-grad-err_2} to the second term on the right-hand side, and the identity $\sum_{i=1}^{N}\|\nabla_{{\theta_i}} F(\lambda^{\pi_{\theta^{k-1}}})\|^2 = \|\nabla_{{\theta}} F(\lambda^{\pi_{\theta^{k-1}}})\|^2$. Then, combining \eqref{lm:marl-contraction-sto-2} and \eqref{lm:marl-contraction-sto-3} yields
	\begin{align}
		\label{lm:marl-contraction-sto-4}
		\|\bar w^{k+1}-w^{k+1}_*\|^2 
		\leq&  \frac{2C_w^2(\eta_\theta^{k-1})^2}{\eta_w^k\mu_w}\Big(\|\nabla_{{\theta}} F(\lambda^{\pi_{\theta^{k-1}}})\|^2 + \zeta_\theta^{k-1} +6E_{\theta^{k-1}}^2+\cO( \gamma^{2H_{k-1}}) \\
		&+ \frac{6C_\phi^2C_\pi^2}{(1-\gamma)^2}\!\sum_{i=1}^{N}\!\left(\!\|\bar w^{k}-w^{k}_*\|^2+\| w^{k}_i-\bar w^{k}\|^2\!\right)\!\!\Big) + \left(\!1-\frac{\eta_w^k\mu_w}{2}\!\right)\!\|\bar w^k \!-\! w^{k}_* \|^2 \!+\! \frac{2\eta_w^k\|\zeta_W^k\|_F^2}{\mu_w\cdot N} \nonumber\\
		\leq& \!  \left(\!\!1\!-\!\frac{\eta_w^k\mu_w}{2} \!+\! \frac{2C_w^2(\eta_\theta^{k-1})^2}{\eta_w^k}\cdot\frac{6NC_\phi^2C_\pi^2}{(1-\gamma)^2}\!\right)\!\cdot\!\|\bar w^k\! -\! w^{k}_* \|^2 \!+\! \frac{2C_w^2(\eta_\theta^{k-1})^2}{\eta_w^k\mu_w}\|\nabla_{{\theta}} F(\!\lambda^{\pi_{\theta^{k-1}}}\!)\!\|^2\nonumber \\
		& + \frac{2\eta_w^k\|\zeta_W^k\|_F^2}{\mu_w\cdot N}+ \frac{2C_w^2(\eta_\theta^{k-1})^2}{\eta_w^k\mu_w}(\zeta_\theta^{k-1} +6E_{\theta^{k-1}}^2+\cO( \gamma^{2H_{k-1}}))\nonumber\\
		&\quad + \frac{2C_w^2(\eta_\theta^{k-1})^2}{\eta_w^k\mu_w}\cdot\frac{6C_\phi^2C_\pi^2}{(1-\gamma)^2}\cdot\sum_{i=1}^{N}\| w^{k}_i-\bar w^{k}\|^2\nonumber
	\end{align} 
	Because the step-size is selected such that $\eta_\theta^{k-1}\leq\frac{(1-\gamma)\mu_w\eta_w^k}{4\sqrt{3N}C_wC_\phi C_\pi},$	then	$\frac{2C_w^2(\eta_\theta^{k-1})^2}{\eta_w^k\mu_w}\cdot\frac{6NC_\phi^2C_\pi^2}{(1-\gamma)^2}\leq \frac{\eta_w^k\mu_w}{4}.$
	Consequently, the right-hand side of the preceding expression may be simplified as
	\begin{align*} 
		&\|\bar w^{k+1}-w^{k+1}_*\|^2 \\
		&\leq   \left(1-\frac{\eta_w^k\mu_w}{4} \right)\cdot\|\bar w^k - w^{k}_* \|^2 + \frac{2C_w^2(\eta_\theta^{k-1})^2}{\eta_w^k\mu_w}\|\nabla_{{\theta}} F(\lambda^{\pi_{\theta^{k-1}}})\|^2 + \frac{2\eta_w^k\|\zeta_W^k\|_F^2}{\mu_w\cdot N}\\
		& \quad\!+\! \frac{2C_w^2(\eta_\theta^{k-1})^2}{\eta_w^k\mu_w}\big(\zeta_\theta^{k-1}\!+\!6E_{k\!-\!1}^2\!+\!\cO( \gamma^{2H_{k\!-\!1}})\!\big) \!+\! \frac{2C_w^2(\eta_\theta^{k-1})^2}{\eta_w^k\mu_w}\!\cdot\!\frac{6C_\phi^2C_\pi^2}{(1-\gamma)^2}\!\cdot\!\sum_{i=1}^{N}\| w^{k}_i\!-\!\bar w^{k}\|^2.
	\end{align*}  
	Note that we use $w_*^{k+1}$ and $w_*^k$ to denote $w^*(\theta^k)$ and $w^*(\theta^{k-1})$ respectively, this proves  Lemma \ref{lemma:marl-contraction-sto}.
\end{proof}

\section{Proof of Lemma \ref{lemma:cvg-sto-2ts}}
Before presenting the proof, let us first provide a supporting lemma. 
\begin{lemma}
	\label{lemma:marl-potential-function-sto}
	For Algorithm \ref{alg:stochastic-batch-2ts}, if for any $k\geq0$ we choose the step sizes to be  $\eta_w^{k+1}\leq\eta_w^{k}\leq 1/L_w$, and 
	$\eta_\theta^{k} =  \min\left\{\frac{(1-\gamma)\mu_w\eta_w^{k+1}}{C_wC_\phi C_\pi}\cdot\frac{1}{\max\{4\sqrt{3N},6\sqrt{10}\}}, \frac{1}{4L_\theta}\right\},$
	then it holds that
	\begin{align} 
		\label{lm:marl-potential-function-sto-0}
		&\quad\frac{\eta_\theta^k}{4} \left\|\nabla_{\theta}F(\lambda^{\pi_{\theta^{k}}})\right\|^2 - \frac{\eta_\theta^{k-1}}{8} \left\|\nabla_{\theta}F(\lambda^{\pi_{\theta^{k-1}}})\right\|^2\\
		& \leq  R_{k+1} - R_{k}  +  \frac{9C_\phi^2C_\pi^2\eta_\theta^k}{2(1-\gamma)^2}\cdot\sum_{i=1}^{N}\|\bar w^{k+1} - w^{k+1}_i\|^2+  \frac{3C_\phi^2C_\pi^2\eta_\theta^{k-1}}{4(1-\gamma)^2}\cdot\sum_{i=1}^{N}\|\bar w^{k} - w^{k}_i\|^2\nonumber\\
		&\quad\!+\!\frac{\eta_\theta^{k\!-\!1}}{8}\!\cdot\!\frac{(\eta_w^k)^2\|\zeta_W^k\|_F^2}{NC_w^2(\eta_\theta^{k-1})^2}\!+\!\frac{3\eta_\theta^k}{4}(\zeta_\theta^k \!+\! 6E_{\theta^{k}}^2 \!+\! \cO(\gamma^{2H_k})) \!+\!\frac{\eta_\theta^{k-1}}{8}(\zeta_\theta^{k-1} \!+\! 6E_{\theta^{k-1}}^2\!+\! \cO(\gamma^{2H_{k-1}})).\nonumber
	\end{align}
\end{lemma}
The proof of Lemma \ref{lemma:marl-potential-function-sto} is provided in Appendix \ref{appdx:lemma-marl-potential-function-sto}.

\begin{proof}
	Begin by considering the result of Lemma \ref{lemma:marl-potential-function-sto} and sum over iteration  $k = 1,2,...,T$ :
	\begin{eqnarray} \label{eq:theorem_proof_start}
		&& \sum_{k=1}^{T}\frac{\eta_\theta^k}{8}\left\|\nabla_{\theta}F(\lambda^{\pi_{\theta^{k}}})\right\|^2 - \frac{\eta_\theta^0}{8} \left\|\nabla_{\theta}F(\lambda^{\pi_{\theta^{0}}})\right\|^2 +  \frac{\eta_\theta^T}{8} \left\|\nabla_{\theta}F(\lambda^{\pi_{\theta^{T}}})\right\|^2\\
		& \leq & R_{T+1} - R_{1}  +  \frac{11C_\phi^2C_\pi^2}{2(1-\gamma)^2}\cdot\sum_{k=1}^T\eta_\theta^{k-1}\sum_{i=1}^{N}\|\bar w^{k} - w^{k}_i\|^2+\sum_{k=1}^T\frac{\eta_\theta^{k-1}}{8}\cdot\frac{(\eta_w^{k})^2\|\zeta_W^k\|_F^2}{NC_w^2(\eta_\theta^{k-1})^2}\nonumber\\
		&&  +\sum_{k=0}^T\frac{7\eta_\theta^k}{8}(\zeta_\theta^k + 6E_{\theta^{k}}^2 +\cO(\gamma^{2H_k})) .\nonumber
	\end{eqnarray}
	Consequently, by rearranging the preceding expression such that $\left\|\nabla_{\theta}F(\lambda^{\pi_{\theta^{k}}})\right\|^2$ is on the left-hand side,  and lower-bounding the average gradient-norm-squared by its minimum over  $k$ permits us to write:
	\begin{eqnarray} 
		\label{thm:pf-1}
		&&\frac{\sum_{k=1}^T\eta_\theta^k\left\|\nabla_{\theta}F(\lambda^{\pi_{\theta^{k}}})\right\|^2}{\sum_{k=1}^T\eta_\theta^k}\\ 
		& \leq & \frac{8(R_{T+1} - R_{1})}{\sum_{k=1}^{T}\eta_\theta^k}  + \frac{\eta_\theta^0\left\|\nabla_{\theta}F(\lambda^{\pi_{\theta^{0}}})\right\|^2}{\sum_{k=1}^{T}\eta_\theta^k} +  \frac{44C_\phi^2C_\pi^2}{(1-\gamma)^2}\cdot\frac{\sum_{k=1}^T\eta_\theta^k\sum_{i=1}^{N}\|\bar w^{k} - w^{k}_i\|^2}{\sum_{k=1}^{T}\eta_\theta^k}\nonumber\\
		&&  +\frac{1}{\sum_{k=1}^{T}\eta_\theta^k}\cdot\sum_{k=1}^T\eta_\theta^{k-1}\frac{(\eta_w^k)^2\|\zeta_W^k\|_F^2}{NC_w^2(\eta_\theta^{k})^2}+\frac{7\sum_{k=0}^T\eta_\theta^k(\zeta_\theta^k + 6E_{\theta^{k}}^2+\cO(\gamma^{2H_k}))}{\sum_{k=1}^{T}\eta_\theta^k}.\nonumber
	\end{eqnarray}
	Then Lemma \ref{lemma:stochastic-grad-err} and the union bound indicates that with probability at least $1-3N\sum_{k=0}^{T}\delta_k$, it holds for any $0\leq k\leq T$ that 
	\begin{equation}\label{eq:error_bound_theorem}
		\big\|\zeta_W^k\big\|^2_F \leq \cO\left( \frac{\log(1/\delta_k)C_\phi^2}{(1-\gamma)^2B_k}\Big(C_\phi^2\sum_{i=1}^N\|w_i^k\|^2 + \frac{NC_F^2}{(1-\gamma)^2}+NL_\lambda^2\Big) +  \gamma^{2H_k}\right)
	\end{equation}
	and 
	$$\zeta_\theta^k \leq \cO\Big(\frac{C_\phi^2C_\pi^2}{(1-\gamma)^2}\cdot \frac{N\log(1/\delta_k)\|w_*^{k\!+\!1}\|^2}{B_k}\Big).$$
	Consequently, for \eqref{thm:pf-1}, the term involving the directional error $\zeta_W^k$ satisfies 
	\begin{eqnarray}
		&&\frac{1}{\sum_{k=1}^{T}\eta_\theta^k}\cdot\sum_{k=1}^T\eta_\theta^{k-1}\frac{(\eta_w^k)^2\|\zeta_W^k\|_F^2}{NC_w^2(\eta_\theta^k)^2}\nonumber\\ 
		&\overset{(i)}{\leq}& \frac{\max\{48N,360\}\cdot C_\phi^2C_\pi^2}{N(1-\gamma)^2\mu_w^2}\cdot \frac{\sum_{k=1}^T\eta_\theta^{k-1}\big\|\zeta_W^k\big\|^2_F}{\sum_{k=1}^{T}\eta_\theta^k}\nonumber\\
		&\leq& \frac{C_\phi^2C_\pi^2}{(1-\gamma)^2\mu_w^2}\cdot\frac{1}{\sum_{k=1}^{T}\eta_\theta^k}\sum_{k=1}^{T}\left(\eta_\theta^{k-1}\cO\Big( \frac{\log(1/\delta_k)C_\phi^2}{(1-\gamma)^2B_k}\Big(C_\phi^2\sum_{i=1}^N\|w_i^k\|^2 + \frac{NC_F^2}{(1-\gamma)^2}+NL_\lambda^2\Big) +  \gamma^{2H_k}\Big)\right)\nonumber\\
		&\leq&\cO\left(\frac{\sum_{k=1}^{T}\eta_\theta^{k-1}\cdot\big(\frac{N\log(1/\delta_k)}{B_k}+\gamma^{2H_k}\big)}{\sum_{k=1}^{T}\eta_\theta^k}\right)\nonumber 
	\end{eqnarray}
	where (i) is due to the requirement on $\eta_\theta^k$ and $\eta_w^k$, the second equality makes use of \eqref{eq:error_bound_theorem}, and the last suppresses dependence on complicated constants. Returning to the consensus error term on the right-hand side of \eqref{eq:theorem_proof_start}:
	\begin{eqnarray}\label{eq:directional_error_consensus}
		&&\frac{44C_\phi^2C_\pi^2}{(1-\gamma)^2}\cdot\frac{\sum_{k=1}^T\eta_\theta^k\sum_{i=1}^{N}\|\bar w^{k} - w^{k}_i\|^2}{\sum_{k=1}^{T}\eta_\theta^k}\\
		&\overset{(i)}{\leq}&\frac{44C_\phi^2C_\pi^2}{(1-\gamma)^2}\cdot\frac{\sum_{k=1}^T2\eta_\theta^k\left(\max_{k'\leq k}\|\zeta_W^{k'}\|_F^2 + \frac{NC_F^2C_\phi^2}{(1-\gamma)^4}\right)\cdot\left(\sum_{k'=0}^k\eta_w^{k'}\rho^{m(k-k')}\right)^2\cdot\rho^{2m}}{\sum_{k=1}^{T}\eta_\theta^k}\nonumber\\ 
		&\overset{(ii)}{\leq}& \cO\left(\frac{NC_\phi^4C_F^2C_\pi^2\rho^{2m}}{(1-\gamma)^6}\cdot\frac{\sum_{k=1}^T\eta_\theta^k\cdot\left(\sum_{k'=0}^k\eta_w^{k'}\rho^{m(k-k')}\right)^2}{\sum_{k=1}^{T}\eta_\theta^k}\right)\nonumber\\
		& = & \cO\left(\frac{\sum_{k=1}^T\eta_\theta^k\cdot\left(\sum_{k'=0}^k\eta_w^{k'}\rho^{m(k-k')}\right)^2}{\sum_{k=1}^{T}\eta_\theta^k}\cdot\rho^{2m}\right)\nonumber
	\end{eqnarray}
	where (i) is due to Lemma \ref{lemma:marl-dis-from-avg-sto} and (ii) follows via our choice of batch size $B_k$ and horizon length $H_k$ horizon yielding the condition $\|\zeta_W^k\|_F^2  = \cO(B_k^{-1} + \gamma^{2H_k})\leq \cO\left(\frac{NC_F^2C_\phi^2}{(1-\gamma)^4}\right)$. 
	Now we study the directional error of the actor $ \zeta_\theta^k$ on the right-hand side of \eqref{eq:theorem_proof_start}:
	\begin{align}\label{eq:directional_error_actor}
		\frac{7\sum_{k=0}^T\eta_\theta^k\zeta_\theta^k}{\sum_{k=1}^{T}\eta_\theta^k} \leq& \cO\left(\frac{NC_\phi^2C_\pi^2}{(1-\gamma)^2}\cdot\frac{\sum_{k=1}^{T} \eta_\theta^k\frac{\log(1/\delta_k)\|w_*^{k\!+\!1}\|^2}{B_k}}{\sum_{k=1}^{T}\eta_\theta^k}\right) 	\nonumber
		\\
		=& \cO\left(\frac{\sum_{k=1}^{T} \eta_\theta^k\cdot\log(1/\delta_k)\cdot B_k^{-1}}{\sum_{k=1}^{T}\eta_\theta^k}\right).
	\end{align}
	Combining \eqref{thm:pf-1} - \eqref{eq:directional_error_actor} then allows us to establish the result. 
\end{proof}

\section{Proof of Lemma \ref{lemma:marl-potential-function-sto}}
\label{appdx:lemma-marl-potential-function-sto}
\begin{proof}
	Begin with a rearrangement of the expression in Lemma \ref{lemma:sufficient-ascent} in terms of an upper-bound on the gradient-norm:
	\begin{eqnarray}
		\label{lm:marl-potential-function-sto-1}
		\frac{\eta_\theta^k}{4} \left\|\nabla_{\theta}F(\lambda^{\pi_{\theta^{k}}})\right\|^2
		&\leq& F(\lambda^{\pi_{\theta^{k+1}}}) - F(\lambda^{\pi_{\theta^{k}}}) + \frac{3\eta_\theta^k}{4}\sum_{i=1}^{N}\big\|\nabla_{{\theta_i}}F(\lambda^{\pi_{\theta^{k}}})-\widehat{\Delta}_{\theta_{i}}^k\big\|^2\\
		&\overset{(i)}{\leq}& F(\lambda^{\pi_{\theta^{k+1}}}) - F(\lambda^{\pi_{\theta^{k}}}) +  \frac{9\eta_\theta^k C_\phi^2C_\pi^2}{4(1-\gamma)^2}\sum_{i=1}^{N}\|w^{k+1}_i-w^{k+1}_*\|^2+\frac{3\eta_\theta^k}{4}\left(\zeta_\theta^k + 6E_{\theta^{k}}^2 + \cO(\gamma^{2H_k})\right)\nonumber\\
		& \leq & F(\lambda^{\pi_{\theta^{k+1}}}) - F(\lambda^{\pi_{\theta^{k}}})  +\frac{3\eta_\theta^k}{4}\left(\zeta_\theta^k + 6E_{\theta^{k}}^2 + \cO(\gamma^{2H_k})\right)\nonumber\\
		&&+\frac{9C_\phi^2C_\pi^2\eta_\theta^k}{2(1-\gamma)^2}\sum_{i=1}^{N}\left(\|w^{k+1}_i-\bar w^{k+1}\|^2 + \|\bar w^{k+1} - w^{k+1}_*\|^2\right)\nonumber\\
		& \overset{(ii)}{=} & R_{k+1} - R_{k}  +\frac{3\eta_\theta^k}{4}\left(\zeta_\theta^k + 6E_{\theta^{k}}^2 + \cO(\gamma^{2H_k})\right) + \frac{9C_\phi^2C_\pi^2\eta_\theta^k}{2(1-\gamma)^2}\sum_{i=1}^{N}\|w^{k+1}_i-\bar w^{k+1}\|^2\nonumber\\
		& & - \alpha\sum_{i=1}^{N}\|\bar w^{k} - w^{k}_*\|^2 + \left(\frac{9C_\phi^2C_\pi^2\eta_\theta^k}{2(1-\gamma)^2}+\alpha\right)\sum_{i=1}^{N}\|\bar w^{k+1} - w^{k+1}_*\|^2\nonumber
	\end{eqnarray}
	where (i) is due to Lemma \ref{lemma:stochastic-grad-err}\ref{lemma:stochastic-grad-err_2}, (ii) is due the definition of the potential function \eqref{defn:potential}. 
	Now apply Lemma \ref{lemma:marl-contraction-sto} to the last term on the right-hand side of the preceding expression to obtain
	\begin{align*}
		&\qquad\,\,\big(\frac{9C_\phi^2C_\pi^2\eta_\theta^k}{2(1-\gamma)^2}+\alpha\big)\sum_{i=1}^{N}\|\bar w^{k+1} - w^{k+1}_*\|^2 - \alpha\sum_{i=1}^{N}\|\bar w^{k} - w^{k}_*\|^2\\
		& \qquad\quad\leq  \left(\frac{9C_\phi^2C_\pi^2\eta_\theta^k}{2(1-\gamma)^2}+\alpha\right)\cdot\Bigg( \frac{2C_w^2(\eta_\theta^{k-1})^2}{\eta_w^k\mu_w}\|\nabla_{{\theta}} F(\lambda^{\pi_{\theta^{k-1}}})\|^2 + \frac{2C_w^2(\eta_\theta^{k-1})^2}{\eta_w^k\mu_w}\Big(\zeta_\theta^{k-1}  \\
		&\qquad \qquad+ 6E_{\theta^{k-1}}^2+ \cO(\gamma^{2H_{k-1}})\Big)+ \frac{2\eta_w^k\|\zeta_W^k\|_F^2}{\mu_w\cdot N}+ \frac{2C_w^2(\eta_\theta^{k-1})^2}{\eta_w^k\mu_w}\cdot\frac{6C_\phi^2C_\pi^2}{(1-\gamma)^2}\cdot\sum_{i=1}^{N}\|\bar w^{k} - w^{k}_i\|^2 \Bigg)\nonumber\\
		& \qquad\qquad+ \left\{\left(1-\frac{\eta_w^k\mu_w}{4}\right)\left(\frac{3C_\phi^2C_\pi^2\eta_\theta^k}{(1-\gamma)^2}+\alpha\right) - \alpha\right\}\sum_{i=1}^{N}\|\bar w^{k} - w^{k}_*\|^2 
	\end{align*}
	Now select the coefficient as 
	$$\alpha =  \frac{18C_\phi^2C_\pi^2}{(1-\gamma)^2\mu_w}\cdot\max_{k\geq 0}\left\{\frac{\eta_\theta^k}{\eta_w^k}\right\}$$ 
	in the potential function $R_k$ [cf. \eqref{defn:potential}] such that the coefficient before $\sum_{i=1}^{N}\|\bar w^{k} - w^{k}_*\|^2$ is nonpositive. Doing so then yields
	\begin{eqnarray}
		\label{lm:marl-potential-function-sto-2}
		&&\left(\frac{9C_\phi^2C_\pi^2\eta_\theta^k}{2(1-\gamma)^2}+\alpha\right)\sum_{i=1}^{N}\|\bar w^{k+1} - w^{k+1}_*\|^2 - \alpha\sum_{i=1}^{N}\|\bar w^{k} - w^{k}_*\|^2\\
		&\overset{(i)}{\leq}&\left(\frac{9C_\phi^2C_\pi^2\eta_\theta^k}{2(1-\gamma)^2\mu_w\eta_w^k}+\alpha\right)\sum_{i=1}^{N}\|\bar w^{k+1} - w^{k+1}_*\|^2 - \alpha\sum_{i=1}^{N}\|\bar w^{k} - w^{k}_*\|^2\nonumber\\
		& \leq & \frac{45C_\phi^2C_\pi^2}{2(1-\gamma)^2\mu_w}\cdot\max_{k'\geq 0}\left\{\frac{\eta_\theta^{k'}}{\eta_w^{k'}}\right\}\cdot\Bigg( \frac{2C_w^2(\eta_\theta^{k-1})^2}{\eta_w^k\mu_w}\|\nabla_{{\theta}} F(\lambda^{\pi_{\theta^{k-1}}})\|^2 + \frac{2C_w^2(\eta_\theta^{k-1})^2}{\eta_w^k\mu_w}\Big(\zeta_\theta^{k-1} \nonumber\\
		&& + 6E_{\theta^{k-1}}^2 + \cO(\gamma^{2H_{k-1}})\Big)  + \frac{2\eta_w^k\|\zeta_W^k\|_F^2}{\mu_w\cdot N}+ \frac{2C_w^2(\eta_\theta^{k-1})^2}{\eta_w^k\mu_w}\!\cdot\!\frac{6C_\phi^2C_\pi^2}{(1-\gamma)^2}\!\cdot\!\sum_{i=1}^{N}\|\bar w^{k} - w^{k}_i\|^2\Bigg)\nonumber\\
		&\leq & \frac{45C_\phi^2C_\pi^2C_w^2(\eta_\theta^{k-1})^2}{(1-\gamma)^2\eta_w^k\mu_w^2}\cdot\max_{k'\geq 0}\left\{\frac{\eta_\theta^{k'}}{\eta_w^{k'}}\right\}\cdot\Bigg(\|\nabla_{{\theta}} F(\lambda^{\pi_{\theta^{k-1}}})\|^2  + \zeta_\theta^{k-1} + 6E_{\theta^{k-1}}^2 + \cO(\gamma^{2H_{k-1}})  \nonumber\\
		&&\qquad\qquad\qquad\qquad\qquad\qquad\quad\qquad + \frac{(\eta_w^k)^2\|\zeta_W^k\|_F^2}{NC_w^2(\eta_\theta^{k-1})^2} + \frac{6C_\phi^2C_\pi^2}{(1-\gamma)^2}\cdot\sum_{i=1}^{N}\|\bar w^{k} - w^{k}_i\|^2\Bigg)\nonumber
	\end{eqnarray}
	where (i) is because we choose the stepsize $\eta_w^k$ s.t. $\mu_w\eta_w^k\leq\mu_w/L_w\leq1$. Note that the actor update step-size is chosen to satisfy $\eta_\theta^{k}\leq\eta_\theta^{k-1}\leq\frac{(1-\gamma)\mu_w\eta_w^k}{6\sqrt{10}C_wC_\phi C_\pi}$, which implies 
	$\frac{45C_\phi^2C_\pi^2C_w^2(\eta_\theta^{k-1})^2}{(1-\gamma)^2\eta_w^k\mu_w^2}\cdot\max_{k'\geq 0}\left\{\frac{\eta_\theta^{k'}}{\eta_w^{k'}}\right\}\leq \frac{\eta_\theta^{k-1}}{8}$. Under this selection, by combining \eqref{lm:marl-potential-function-sto-1} and \eqref{lm:marl-potential-function-sto-2} we may write
	\begin{align*} 
		&\quad\frac{\eta_\theta^k}{4} \left\|\nabla_{\theta}F(\lambda^{\pi_{\theta^{k}}})\right\|^2 - \frac{\eta_\theta^{k-1}}{8} \left\|\nabla_{\theta}F(\lambda^{\pi_{\theta^{k-1}}})\right\|^2\\
		& \leq  R_{k+1} - R_{k}  +  \frac{9C_\phi^2C_\pi^2\eta_\theta^k}{2(1-\gamma)^2}\cdot\sum_{i=1}^{N}\|\bar w^{k+1} - w^{k+1}_i\|^2+  \frac{3C_\phi^2C_\pi^2\eta_\theta^{k-1}}{4(1-\gamma)^2}\cdot\sum_{i=1}^{N}\|\bar w^{k} - w^{k}_i\|^2\\
		&\quad\!+\!\frac{\eta_\theta^{k\!-\!1}}{8}\!\cdot\!\frac{(\eta_w^k)^2\|\zeta_W^k\|_F^2}{NC_w^2(\eta_\theta^{k-1})^2}\!+\!\frac{3\eta_\theta^k}{4}(\zeta_\theta^k \!+\! 6E_{\theta^{k}}^2 \!+\! \cO(\gamma^{2H_k})) \!+\!\frac{\eta_\theta^{k-1}}{8}(\zeta_\theta^{k-1} \!+\! 6E_{\theta^{k-1}}^2\!+\! \cO(\gamma^{2H_{k-1}})).
	\end{align*}
	This proves the result. 
\end{proof}

\section{Proof of Theorem \ref{theorem:final}}\label{appendix_theorem}
\begin{proof}
	Lemma \ref{lemma:cvg-sto-2ts} says
	\begin{align} 
		\frac{\sum_{k=1}^T\eta_\theta^k\left\|\nabla_{\theta}F(\lambda^{\pi_{\theta^{k}}})\right\|^2}{\sum_{k=1}^T\eta_\theta^k} 
		\leq &\cO\left(\frac{(R_{T+1} - R_{1})+\eta_\theta^0\left\|\nabla_{\theta}F(\lambda^{\pi_{\theta^{0}}})\right\|^2}{\sum_{k=1}^{T}\eta_\theta^k}   \right) +\cO\left(\frac{\sum_{k=0}^{T}\eta_\theta^kE_{\theta^{k}}^2}{\sum_{k=1}^{T}\eta_\theta^k}   \right)\nonumber\\
		&+ \cO\left(\frac{\sum_{k=1}^{T}\eta_\theta^{k-1}\cdot\big(\frac{N\log(1/\delta_k)}{B_k}+\gamma^{2H_k}\big)}{\sum_{k=1}^{T}\eta_\theta^k}\right)\nonumber\\
		&\quad+ \cO\left(\frac{\sum_{k=1}^T\eta_\theta^k\cdot\left(\sum_{k'=0}^k\eta_w^{k'}\rho^{m(k-k')}\right)^2}{\sum_{k=1}^{T}\eta_\theta^k}\cdot\rho^{2m}\right)\nonumber.
	\end{align}
	In the first case where we take constant stepsizes and batchsizes, we have can simplify this inequality as 
	\begin{eqnarray} 
		&&\frac{1}{T}\sum_{k=1}^T\left\|\nabla_{\theta}F(\lambda^{\pi_{\theta^{k}}})\right\|^2
		\leq \cO\left(\frac{1}{T\eta_\theta}+\frac{\sum_{k=0}^{T}E_{\theta^{k}}^2}{T}    + \frac{1}{B}+\gamma^{2H}+\eta_w^2\rho^{2m}\right)\nonumber.
	\end{eqnarray}
	where we write $B_k\equiv B$, $H_k\equiv H$, $\eta_\theta^k \equiv \eta_\theta$, $\eta_w^k\equiv\eta_w$. Consequently, setting these parameters according to the theorem and using the fact that $E_{\theta^k}^2\leq W$ proves that
	$$\frac{1}{T}\sum_{k=1}^T\left\|\nabla_{\theta}F(\lambda^{\pi_{\theta^{k}}})\right\|^2
	\leq \cO\left(\epsilon+W\right), \quad w.p. \quad 1-\delta.$$ 
	The total sample complexity is $T\times B\times H = \tilde{\cO}(\epsilon^{-2.5})$.
	
	For the second case, we should notice that for the consensus error term, for sufficiently large $k$, we have 
	\begin{eqnarray*} 
		\sum_{k'=0}^k\eta_w^{k'}\rho^{m(k-k')} &=& \cO\left(\sum_{k'=0}^k (k'+1)^{-1/3}\rho^{m(k-k')}\right)\\
		&\leq& \cO\left(\sum_{k'=0}^{\lceil k/2\rceil} \rho^{m(k-k')} +  \sum_{k'=\lceil k/2\rceil+1}^{k} (\lceil k/2\rceil+1)^{-1/3}\rho^{m(k-k')}\right)\\
		&=& \cO\left(\rho^{m\lceil k/2\rceil} + (\lceil k/2\rceil+1)^{-1/3}\right)\\
		& = & \cO\left((k+1)^{-1/3}\right).
	\end{eqnarray*}
	As a result, according to our selection of the parameters, we have 
	$\sum_{k=1}^{T}\eta_\theta^k = \Theta(T^{2/3}).$ Notice that $\sum_{k=0}^{T} (k+1)^{-1} = \cO(\log T)$, then choosing $H_k = \cO(\log_\gamma((k+1)^{-2/3})) = \cO((1-\gamma)^{-1}\log(k+1))$ and $B_k = \log(1/\delta_k)(k+1)^{2/3}$ gives
	$$\frac{\sum_{k=1}^{T}\eta_\theta^{k-1}\cdot\big(\frac{N\log(1/\delta_k)}{B_k}+\gamma^{2H_k}\big)}{\sum_{k=1}^{T}\eta_\theta^k} = \cO(T^{-2/3}\log T)$$
	and 
	$$\frac{\sum_{k=1}^T\eta_\theta^k\cdot\left(\sum_{k'=0}^k\eta_w^{k'}\rho^{m(k-k')}\right)^2}{\sum_{k=1}^{T}\eta_\theta^k}\cdot\rho^{2m} = \cO(T^{-2/3}\log T).$$
	substitute the above inequalities into Lemma \ref{lemma:cvg-sto-2ts} proves that
	$$\frac{\sum_{k=1}^T\eta_\theta^k\left\|\nabla_{\theta}F(\lambda^{\pi_{\theta^{k}}})\right\|^2}{\sum_{k=1}^T\eta_\theta^k} 
	\leq \cO\left(\frac{\log T}{T^{\frac{2}{3}}}+W\right) , \quad w.p. \quad 1-\delta,$$
	where the failure probability is due to the following argument
	$$3N\sum_{k=0}^{T}\delta_k\leq 3N\sum_{k=0}^{\infty}\delta_k = 3N\sum_{k=0}^{\infty}\frac{2\delta}{N\pi^2(k+1)^2} = \delta.$$
	To make $\frac{\log T}{T^{\frac{2}{3}}} = \epsilon$, we need $T = \tilde{\cO}(\epsilon^{-3/2})$. Consequently, the total sample complexity will be 
	$$\sum_{k=0}^TB_k\times H_k = \tilde{\cO}\Big(\sum_{k=1}^{\epsilon^{-3/2}}k^{2/3}\Big) = \tilde{\cO}(\epsilon^{-5/2})$$.
\end{proof}
\subsection{Proof of Corollary \ref{corollary:global}}\label{apx_global_optima}
\begin{proof}
	Because $W=0$ in this scenario, the second setting (adaptive parameter selection) of  Theorem \ref{theorem:final} indicates that 
	$$\frac{\sum_{k=1}^T\eta_\theta^k\left\|\nabla_{\theta}F(\lambda^{\pi_{\theta^{k}}})\right\|^2}{\sum_{k=1}^T\eta_\theta^k} 
	\leq \cO\left(\frac{\log T}{T^{\frac{2}{3}}}\right) , \quad w.p. \quad 1-\delta$$
	for any $T$. If we set $\bar\theta^T = \theta^k$ w.p. $\frac{\eta_\theta^k}{\sum_{k'=1}^T\eta_\theta^{k'}}$, $1\leq k\leq T$. Then 
	$$\EE[\|\nabla_{{\theta}} F(\lambda^{\pi_{\bar\theta_T}})\|^2|\cF_T] = \frac{\sum_{k=1}^T\eta_\theta^k\left\|\nabla_{\theta}F(\lambda^{\pi_{\theta^{k}}})\right\|^2}{\sum_{k=1}^T\eta_\theta^k} 
	\leq \cO\left(\frac{\log T}{T^{\frac{2}{3}}}\right) , \quad w.p. \quad 1-\delta.$$
	Therefore, as long as the Assumption 1 of \cite{zhang2020variational} is satisfied, \cite{zhang2020variational} indicates that problem \eqref{prob:main} has no saddle point. In this sense, Algorithm \ref{alg:stochastic-batch-2ts} is converging to the global optima. 
\end{proof}
\subsection{Proof of Corollary \ref{corollary:communication}}\label{apx_commuication}
\begin{proof}
	First, because Corollary \ref{corollary:communication} is also a constant algorithmic parameter setting, the proof of Theorem \ref{theorem:final} indicates that 
	\begin{eqnarray} 
		&&\frac{1}{T}\sum_{k=1}^T\left\|\nabla_{\theta}F(\lambda^{\pi_{\theta^{k}}})\right\|^2
		\leq \cO\left(\frac{1}{T\eta_\theta}+W   + \frac{1}{B}+\gamma^{2H}+\eta_w^2\rho^{2m}\right)\nonumber.
	\end{eqnarray}
	Then choosing $m$ large enough so that $\rho^{2m} = \cO(\epsilon)$ (equivalently $m = \cO((1-\rho)^{-1}\log(\epsilon^{-1}))$) makes $\eta_w^2\rho^{2m} = \cO(\epsilon)$ even if we choose a constant step size for the critic update, i.e., $\eta_w = 1/L_w$. This also enables us to choose a constant step size $\eta_\theta$ for the actor update, as is defined the corollary. Therefore, we only need to set $B = \cO(\epsilon^{-1})$ and $T = \cO(\epsilon^{-1})$, which completes the proof. 	
\end{proof}

\section{Supporting Results}
\label{appdx:support}
\subsection{Existence of the constant $C_w$}
In this section, we prove the following statement in Lemma \ref{lemma:marl-contraction-sto}: $$\exists C_w>0 \quad s.t.\quad \|w_*^{k+1} - w_*^k\|\leq C_w\|\theta^k - \theta^{k-1}\|.$$
\begin{proof}
	First, recall the matrix characterization of the critic problem in \eqref{lm:marl-dis-from-avg-1}, where one has
	\begin{equation*}
		\ell(w;\pi_{\theta^{k}}) = \frac{1}{2}\left\|\Phi^\top w-Q^{\pi_{\theta^{k}}}_F\right\|^2_{\Lambda_k}\quad\mbox{and}\quad\nabla_w\ell(w;\pi_{\theta^{k}}) = \Phi\Lambda_k(\Phi^\top w-Q^{\pi_{\theta^{k}}}_F).
	\end{equation*}
	Denote $w_*^{k+1} = w^*(\theta^k)$ for the ease of notation. Consequently, the KKT condition yields $\nabla_w\ell(w_*^{k+1};\pi_{\theta^{k}}) = \Phi\Lambda_k(\Phi^\top w_*^{k+1}-Q^{\pi_{\theta^{k}}}_F)=0.$ This further indicates that 
	\begin{eqnarray}
		\label{eqn:support-1}
		\|w_*^{k+1}\| & \leq & \|(\Phi\Lambda_k\Phi^\top)^{-1}\|\cdot\| \Phi\Lambda_kQ^{\pi_{\theta^{k}}}_F\|\\
		& \overset{(i)}{\leq} & \mu_w^{-1}\Big\|\sum_{s,a}\lambda^{\pi_{\theta^{k}}}(s,a)\phi(s,a)Q_F^{\pi_{\theta^{k}}}(s,a)\Big\|\nonumber\\
		& \leq & \frac{C_\phi}{\mu_w}\cdot\sum_{s,a}\lambda^{\pi_{\theta^{k}}}(s,a)\cdot|Q_F^{\pi_{\theta^{k}}}(s,a)|\nonumber\\
		& \overset{(ii)}{\leq} &\frac{C_\phi\cdot C_F}{(1-\gamma)^2\cdot\mu_w}\nonumber
	\end{eqnarray} 
	where (i) is because $\Phi\Lambda_k\Phi^\top\succeq\mu_w\cdot I$, and (ii) is because $|Q_F^{\pi_{\theta^{k}}}(s,a)|\leq\frac{C_F}{1-\gamma}$ for any $(s,a)$. 
	
	Next, we show the Lipschitz continuity of the critic solution. Again, by the KKT condition of the critic problem, we have $\Phi\Lambda_k\Phi^\top w_*^{k+1} = \Phi\Lambda_kQ^{\pi_{\theta^{k}}}_F$ and $\Phi\Lambda_{k-1}\Phi^\top w_*^{k} = \Phi\Lambda_{k-1}Q^{\pi_{\theta^{k-1}}}_F$. Consequently, 
	\begin{eqnarray}
		&& \Phi\Lambda_k\Phi^\top (w_*^{k+1} - w_*^{k})\\
		& = & \Phi\Lambda_{k}Q^{\pi_{\theta^{k}}}_F - \Phi\Lambda_k\Phi^\top w_*^{k}\nonumber\\
		& = & \Phi\Lambda_{k}Q^{\pi_{\theta^{k}}}_F - \Phi\Lambda_k\Phi^\top w_*^{k} + (\Phi\Lambda_{k-1}\Phi^\top w_*^{k} - \Phi\Lambda_{k-1}Q^{\pi_{\theta^{k-1}}}_F)\nonumber\\
		& = & \underbrace{\Phi\Lambda_{k}Q^{\pi_{\theta^{k}}}_F-\Phi\Lambda_{k}Q^{\pi_{\theta^{k-1}}}_F}_{T_1}  + \underbrace{\Phi\Lambda_{k}Q^{\pi_{\theta^{k-1}}}_F  - \Phi\Lambda_{k-1}Q^{\pi_{\theta^{k-1}}}_F}_{T_2} + \underbrace{\Phi\Lambda_{k-1}\Phi^\top w_*^{k}- \Phi\Lambda_k\Phi^\top w_*^{k}}_{T_3}\nonumber
	\end{eqnarray} 
	For the first term, 
	\begin{eqnarray*}
		\|T_1\| & = & \Big\|\sum_{s,a}\lambda^{\pi_{\theta^{k}}}(s,a)\phi(s,a)(Q^{\pi_{\theta^{k}}}_F(s,a)-Q^{\pi_{\theta^{k-1}}}_F(s,a))\Big\|\\
		& \leq & \frac{C_\Phi}{1-\gamma}\cdot\|Q^{\pi_{\theta^{k}}}_F-Q^{\pi_{\theta^{k-1}}}_F\|_\infty \\
		& \leq & \frac{C_\Phi\ell_Q}{1-\gamma}\|\theta^{k} - \theta^{k-1}\|,
	\end{eqnarray*}
	\begin{eqnarray*}
		\|T_2\| & = & \Big\|\sum_{s,a}(\lambda^{\pi_{\theta^{k}}}(s,a)-\lambda^{\pi_{\theta^{k-1}}}(s,a))\phi(s,a)Q^{\pi_{\theta^{k-1}}}_F(s,a)\Big\|\\
		& \leq & \frac{C_\Phi C_F}{(1-\gamma)^2}\cdot\|\lambda^{\pi_{\theta^{k}}}-\lambda^{\pi_{\theta^{k-1}}}\|_\infty \\
		& \leq & \frac{C_\Phi C_F\ell_\lambda}{(1-\gamma)^2}\|\theta^{k} - \theta^{k-1}\|,
	\end{eqnarray*}
	\begin{eqnarray*}
		\|T_3\| & \leq & \|\Phi\Lambda_{k-1}\Phi^\top - \Phi\Lambda_k\Phi^\top\|_F\cdot\|w_*^k\|\\
		& \leq & \frac{C_\Phi\cdot C_F}{(1-\gamma)^2\cdot\mu_w}\cdot \Big\|\sum_{s,a}(\lambda^{\pi_{\theta^{k}}}(s,a)-\lambda^{\pi_{\theta^{k-1}}}(s,a))\phi(s,a)\phi(s,a)^\top\Big\|_F\\
		& \leq & \frac{C_\Phi^3\cdot C_F}{(1-\gamma)^2\cdot\mu_w}\cdot \big\|\lambda^{\pi_{\theta^{k}}}-\lambda^{\pi_{\theta^{k-1}}}\big\|_1\\
		& \leq & \frac{|\cS||\cA|C_\Phi^3C_F\ell_\lambda}{(1-\gamma)^2\cdot\mu_w}\cdot \big\|\theta^{k} - \theta^{k-1}\big\|
	\end{eqnarray*}
	Now, combing the above terms with \eqref{eqn:support-1}, we have 
	\begin{eqnarray}
		\|w_*^{k+1} - w_*^k\| & \leq & \|(\Phi\Lambda_{k}\Phi^\top)^{-1}\|(\|T_1\| + \|T_2\| + \|T_3\|)\nonumber\\
		& \leq & \left( \frac{C_\Phi\ell_Q}{(1-\gamma)\cdot\mu_w}+\frac{C_\Phi C_F\ell_\lambda}{(1-\gamma)^2\cdot\mu_w}+\frac{|\cS||\cA|C_\Phi^3C_F\ell_\lambda}{(1-\gamma)^2\cdot\mu_w^2}\right)\cdot \big\|\theta^{k} - \theta^{k-1}\big\|\nonumber\\
		& := & C_w\cdot\big\|\theta^{k} - \theta^{k-1}\big\|.\nonumber
	\end{eqnarray}
\end{proof}

\subsection{Boundedness of critic parameters}
In this section, we prove that as long as the algorithmic parameters are chosen according to Theorem \ref{theorem:final}:
$$\exists D_w>0 \quad s.t.\quad \max_{k\geq 0}\{\|w_*^{k+1}\|, \|w_i^k\|\} \leq D_w, \quad w.p.\quad 1-\delta.$$
\begin{proof}
	First, the proof in the last seciton already indicates that $\|w_*^{k+1}\|$ are all bounded. 
	
	Second, for the other term, 
	by lemma \ref{lemma:marl-contraction-sto}, we have 
	\begin{align}
		\label{eqn:support-2}
		\|\bar w^{k+1}-w^{k+1}_*\|^2 
		& \leq   \left(1-\frac{\eta_w^k\mu_w}{4} \right)\cdot\|\bar w^k - w^{k}_* \|^2 + \frac{2C_w^2(\eta_\theta^{k-1})^2}{\eta_w^k\mu_w}\|\nabla_{{\theta}} F(\lambda^{\pi_{\theta^{k-1}}})\|^2 + \frac{2\eta_w^k\|\zeta_W^k\|_F^2}{\mu_w\cdot N}\\
		&\qquad + \frac{2C_w^2(\eta_\theta^{k-1})^2}{\eta_w^k\mu_w}\left(\zeta_\theta^{k-1} +6E_{\theta^{k-1}}^2+\cO( \gamma^{2H_{k-1}}) + \frac{6C_\phi^2C_\pi^2}{(1-\gamma)^2}\cdot\sum_{i=1}^{N}\| w^{k}_i-\bar w^{k}\|^2\right)\nonumber.
	\end{align}
	It is worth noting that due to our requirement on the stepsize: 
	$$\eta_\theta^{k-1} =  \min\!\Big\{\!\frac{(1\!-\!\gamma)\mu_w\eta_w^{k}}{C_wC_\phi C_\pi}\!\cdot\!\frac{1}{\max\{4\sqrt{3N},6\sqrt{10}\}}, \frac{1}{4L_\theta}\Big\}.$$
	Then 
	$$\frac{(\eta_\theta^{k-1})^2}{\eta_w^k}  = \min\left\{\frac{(1\!-\!\gamma)^2\mu_w^2}{C_w^2C_\phi^2 C_\pi^2}\!\cdot\!\frac{1}{\max\{48N,360\}}, \frac{1}{16L_\theta(\eta_w^k)^2}\right\}\cdot\eta_w^{k}.$$
	Because the step size $\eta_w^k$ is either a constant or diminishing, the above coefficient before $\eta_w^k$ is bounded. Let us denote 
	$c_1 := \sup_{k\geq0} \min\left\{\frac{(1\!-\!\gamma)^2\mu_w^2}{C_w^2C_\phi^2 C_\pi^2}\!\cdot\!\frac{1}{\max\{48N,360\}}, \frac{1}{16L_\theta(\eta_w^k)^2}\right\}$. Therefore, \eqref{eqn:support-2} indicates
	\begin{align}
		\label{eqn:support-3}
		\|\bar w^{k+1}-w^{k+1}_*\|^2 
		& \leq   \left(1-\frac{\eta_w^k\mu_w}{4} \right)\cdot\|\bar w^k - w^{k}_* \|^2 + \frac{\eta_w^k\mu_w}{4}\Bigg(\frac{8c_1C_w^2}{\mu_w^2}\|\nabla_{{\theta}} F(\lambda^{\pi_{\theta^{k-1}}})\|^2+ \frac{8\|\zeta_W^k\|_F^2}{\mu_w^2\cdot N}\\ 
		&\qquad + \frac{8c_1C_w^2}{\mu_w^2}\Big(\zeta_\theta^{k-1} +6E_{\theta^{k-1}}^2+\cO( \gamma^{2H_{k-1}}) + \frac{6C_\phi^2C_\pi^2}{(1-\gamma)^2}\cdot\sum_{i=1}^{N}\| w^{k}_i-\bar w^{k}\|^2\Big)\Bigg)\nonumber.
	\end{align}
	According to Assumption \ref{assumption:Utility}, $\|\nabla_{{\theta}} F(\lambda^{\pi_{\theta^{k-1}}})\|^2$ is bounded. According to Lemma \ref{lemma:stochastic-grad-err}, $\max_{k\geq0}\{\|\zeta_W^k\|_F^2,\zeta_\theta^{k}\}$ is bounded w.p. $1-\delta$. By Assumption \ref{assumption:model-err}, $E_{\theta^{k-1}}^2$ are all bounded by $W$. By Lemma \ref{lemma:marl-dis-from-avg-sto}, $\max_{k\geq 0}\sum_{i=1}^{N}\| w^{k}_i-\bar w^{k}\|^2$ is bounded. Let us denote the upper bound of the term in \eqref{eqn:support-3} inside the bracket with coefficient $\frac{\eta_w^k\mu_w}{4}$ as $c_2$. We have 
	\begin{align}
		\|\bar w^{k+1}-w^{k+1}_*\|^2 
		\leq &   \left(1-\frac{\eta_w^k\mu_w}{4} \right)\cdot\|\bar w^k - w^{k}_* \|^2 + \frac{\eta_w^k\mu_w}{4}c_2\nonumber
		\\
		\leq& \max\{\|\bar w^k - w^{k}_* \|^2, c_2\}\nonumber.
	\end{align}
	Repeat the above inequality yields 
	$$\|\bar w^{k}-w^{k}_*\|^2 
	\leq \max\{\|\bar w^1 - w^{1}_* \|^2, c_2\}, \quad \forall k\geq 1\nonumber.$$
	Also note that Lemma \ref{lemma:marl-dis-from-avg-sto} indicates that $\max_{k\geq 0}\sum_{i=1}^{N}\| w^{k}_i-\bar w^{k}\|^2$ is bounded and previous proof shows that $\|w^{k}_*\|$ is bounded. Therefore, the $\|w^{k}_i\|$ are also bounded for all $k,i$.     
\end{proof}

\section{Additional Experiments and Details}\label{appendix_experiments}
We have performed experiments for single as well as multi agent network for both the \emph{MountainCar} of OpenAI Gym \cite{brockman2016openai} and \emph{Cooperative navigation} environments  \cite{lowe2017multi}.  Since the state space is continuous for both the environments, we use discretization  of the state space to estimate the respective occupancy measure.  Unless otherwise stated, we have used a $2$ layer with $64$ nodes per layer deep neural network (DNN) for the actor as well as critic in the experiments. We use a learning rate of $0.001$ for all the experiments and a batch size of $10$ for the count based density estimator. One epoch in the experiment consists of $1000$ episodes unless otherwise stated and one episode is a one trajectory in the environment. The maximum number of steps per episode are $300$ for the \emph{MountainCar} environment and its $50$ for the \emph{Cooperative navigation} environment. We have reported running averages for all the results reported in this paper, such as, the general utility, the constraint violation, and the consensus error. Next, we provide further details for each set of experiment we presented in the paper.
\begin{figure}[!ht]
	\centering
	\begin{minipage}{\textwidth}
		\centering
		\subfigure[ Entropy comparison]{\includegraphics[scale=0.32]{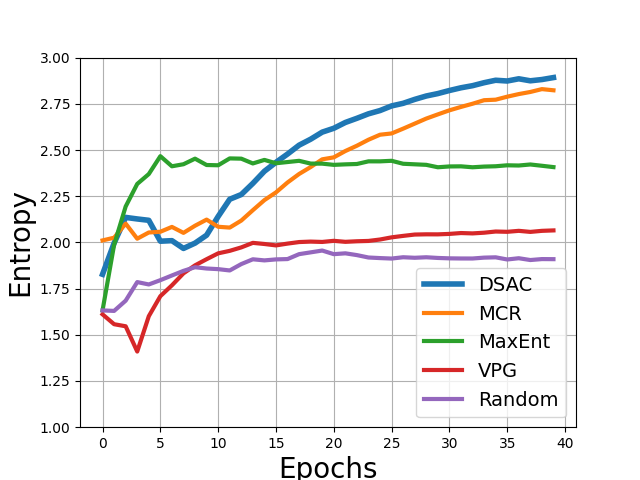}\label{fig1_entropy0}}
		\hspace{0cm}
		\subfigure[ State space coverage]{\includegraphics[scale=0.32]{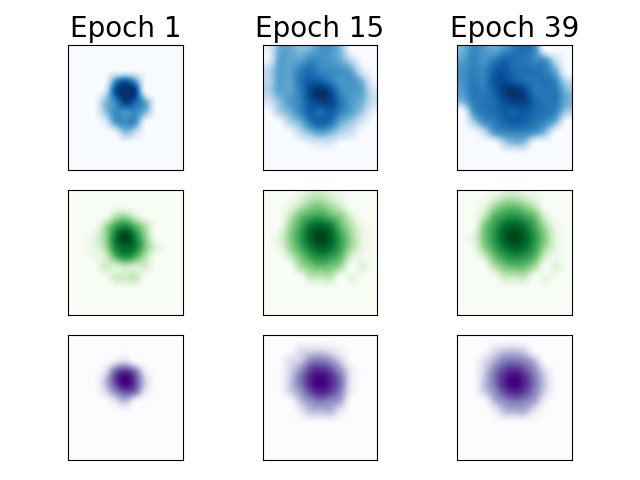}\label{fig1_heatmap}}
		\vspace{-0mm}	\captionof{figure}{  (a) Entropy comparisons, (b) occupancy measure heat
			maps for \emph{MountainCar} environment. Shadow Reward actor-critic
			achieves superior limiting entropy for this instance. }
		\label{fig_single_entropy}
	\end{minipage}\\
	\begin{minipage}{\textwidth}
		\centering
		\subfigure[ World model]{\includegraphics[width=0.25\columnwidth,height=0.13\textheight]{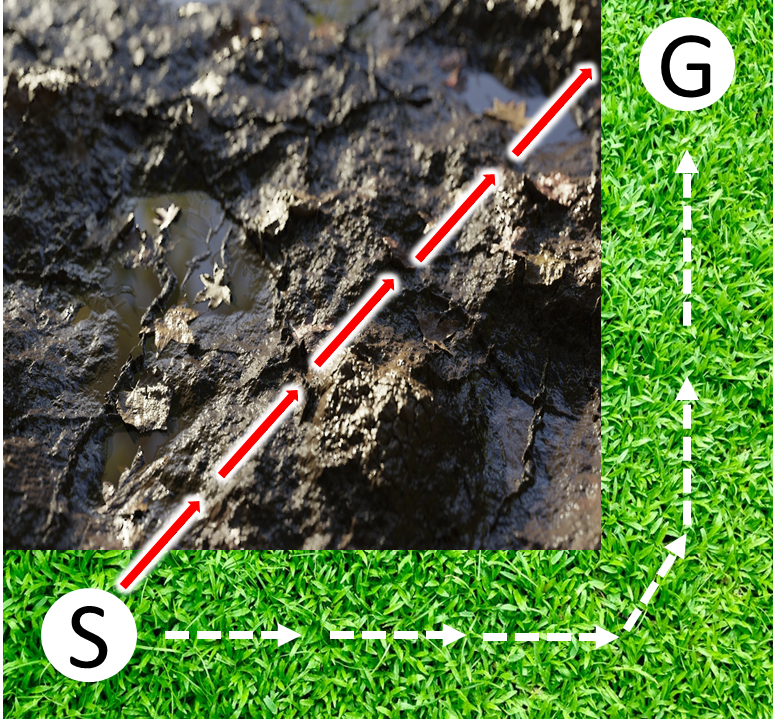}\label{fig2_CMDP_env}}
		\subfigure[ Average return]{\includegraphics[scale=0.30]{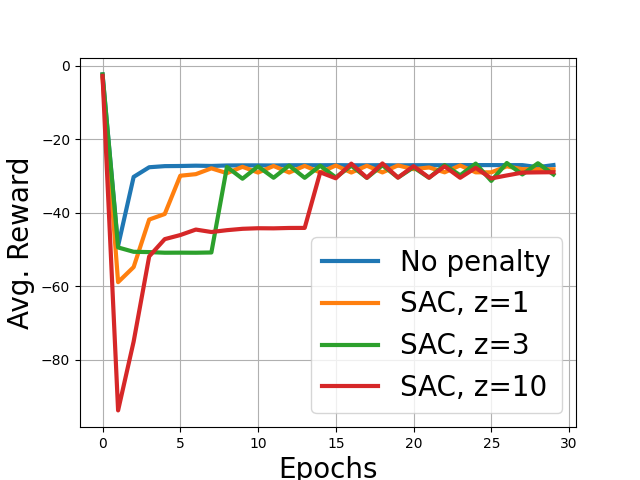}\label{fig2_CMDP_reward}}
		\hspace{0cm}
		\subfigure[ Average cost]{\includegraphics[scale=0.30]{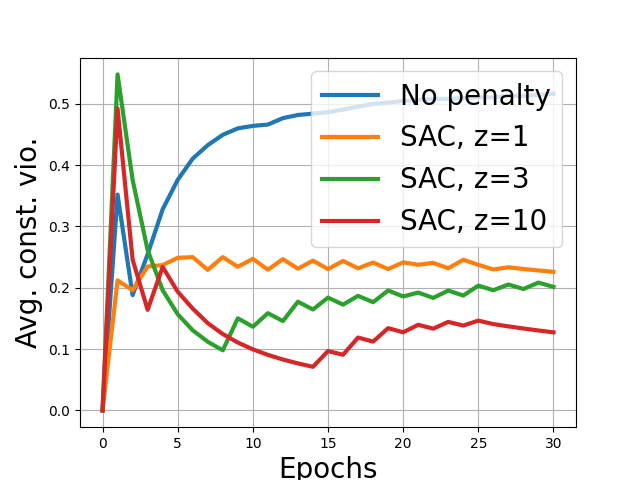}\label{fig2_CMDP_cost}}
		\vspace{-0mm}	\captionof{figure}{ (a) World model for single-agent navigation with green as safe and brown as unsafe
			states. The goal is to travel from Start to Goal safely. (b) Undiscounted average reward return comparison. (c) Average constraint violation comparison for different values of penalty parameter z [cf. \eqref{eq:penalty}]. Constraints yield avoidance of unsafe regions, in contrast to maximizing cumulative return.}
		\label{fig_multi_agent}
	\end{minipage}
	\vspace{-0mm}
\end{figure}

%

\subsection{Single Agent Environment}\label{additional_single}
\textbf{Exploration Maximization.} We solve the problem of maximum entropy exploration for continuous \emph{MountainCar} \cite{brockman2016openai}, where the goal is to maximize the entropy of the long-term occupancy measure $F(\lambda)=-\sum_{s}\lambda_s(\log(\lambda_s))$.  The two dimensional continuous state space is discretized into a $[10,10]$ grid size for occupancy measure estimation. There are three discrete actions available $[\leftarrow, \text{stay},\rightarrow]$. We used the simple count based estimator which counts the state visitation frequency to each bin. We use a fixed batch size of $100$ to estimate the occupancy measure. The actor and critic employ deep neural network (DNN) parametrization as defined by two fully connected hidden layers composed of $128$ nodes each with ReLU activations. 
We compare DSAC with a Monte Carlo rollout (MCR) based estimation for the shadow value function in a manner reminiscent of REINFORCE and a random baseline when agent took actions randomly with uniform distribution. In Fig. \ref{fig1_entropy0}, we have compared the performance with MaxEnt \cite{hazan2018provably} and Variational Policy gradient (VPG) \cite{zhang2020variational} for a neural parameterization of actor and critic. Fig. \ref{fig1_heatmap} shows the heatmap of occupancy measure obtained at three different epochs $1, 15, 39$ for DSAC (blue), MaxEnt (green), Random (purple). Observe that our approaches based on shadow rewards (DSAC and the Monte Carlo variant (MCR)) yield superior performance in terms of entropy and hence space coverage (see Appendix \ref{additional_single} for implementation details).

\textbf{Safe Navigation.} We experiment with a single agent version of \emph{Cooperative navigation} environment  \cite{lowe2017multi}, which is depicted in Fig. \ref{fig2_CMDP_env}: the task is to go from the start to goal without passing through the unsafe (brown) region (green represents the safe region). We impose the safety via constraints $\ip{\lambda^\pi,c} \leq C$ on accumulated costs $c(s_t,a_t)$ via a quadratic penalty:\vspace{-0mm}
\begin{equation}\label{eq:penalty}
	F(\lambda^\pi) = \langle \lambda^\pi, r\rangle - z \left(\ip{\lambda^\pi,c}-C\right)^2,
\end{equation}
where $z$ is the penalty parameter. 
The two-dimensional location of the agent represents the state and five discrete actions are available to agent as $[\leftarrow,\rightarrow,\uparrow,\downarrow,\text{stay}]$. We discretize the space into $[200,200]$ size grid for the occupancy measure estimation. The actor and the critic are two-layer fully-connected DNNs with $64$ nodes at each layer and ReLU activations. 
The performance of the proposed algorithm is shown in Fig.\ref{fig2_CMDP_reward}-\ref{fig2_CMDP_cost} for different values of penalty $z$. Experimentally, we consider a fixed cost of $c(s_t,a_t)=1$ for every visit to the unsafe region with a threshold of $C=0.001$ (see Appendix \ref{additional_single} for details).  Observe that penalization yields behavior that avoids the unsafe region as compared to the unconstrained objective.

\begin{figure}[t]
	\centering{\includegraphics[scale=0.4]{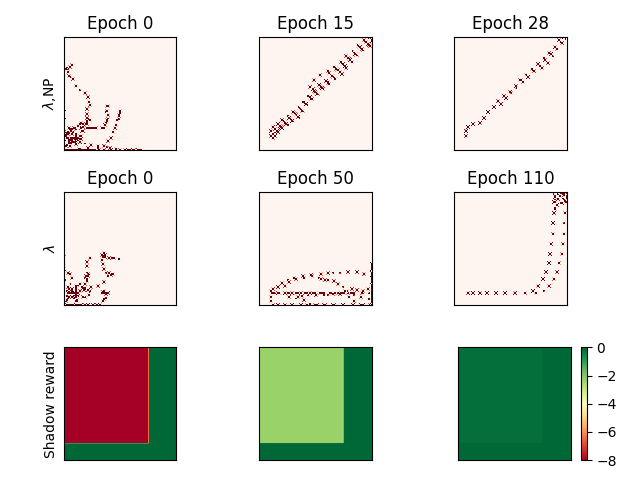}\label{fig1_CMDP}}
	\captionof{figure}{ {Occupancy measure and shadow reward for \emph{safe navigation} environment shown in Fig. \ref{fig_environment_multi}. The heatmaps here show the discretized version of the two dimensional state space (location of the agent) in the environment. The goal is to reach from start (bottom left) to the goal (upper right) while avoiding the mud.The first row depicts that trajectories traverse through the unsafe region since there is no penalization happening for $z=0$ at the outset of training. The second row shows the occupancy measure estimate when we use $z=10$ later on in training, which illuminates that the learned trajectory begins to avoid the unsafe region. The shadow reward over the course of training places a highly negative reward (red) for visiting the unsafe region and zero reward (green) for the safe region, which yields avoidance behavior. The result is depicted in the third plot of second row, which satisfies the safety constraints.} }
	\label{fig_single_agent_entropy22}\vspace{-0mm}
\end{figure}

To illuminate the role of shadow rewards for the safe navigation problem (cf. Fig. \ref{fig2_CMDP_env}), see Fig. \ref{fig_single_agent_entropy22}. The  shadow reward arises from the second penalty term we add to the objective in \eqref{eq:penalty} the appendix. In additional to the reward $r$, we also receive a an additional reward due to the penalty we have in the objective (see \eqref{eq:penalty}). 
We first report the estimated occupancy measure for different epochs in the first row of Fig. \ref{fig_single_agent_entropy22} for $z=0$ (zero penalty case). We note that at epoch $0$, the algorithm samples nearly uniformly across the space of trajectories from start to goal, and has not internalized any penalization associated with unsafe zones. -- see the first row of Fig. \ref{fig_single_agent_entropy22} (epoch $28$). Over the course of training, the shadow reward for all the states incentivizes avoidance of unsafe states as may be observed in the last row of Fig. \ref{fig_single_agent_entropy22} (epoch $0$) with the corresponding occupancy measure shown in the second row. In the last row of Fig. \ref{fig_single_agent_entropy22}, red means the low reward and green means the high reward states. Hence, the proposed algorithm creates a shadow reward which discourages visitation of the unsafe region and hence obtains a trajectory through the safe region as shows in second row of Fig. \ref{fig_single_agent_entropy22} for epoch $28$.

\subsection{Multi Agent Environment}\label{additional_multi}
\textbf{Exploration Maximization.} For the two agent case we are considering as shown in Fig \ref{fig_environment_multi}, the local two-dimensional state of each agent is their position in the $x-y$ plain, and the action space is a discrete set containing five choices $[\leftarrow,\rightarrow,\uparrow,\downarrow,\text{stay}]$. The network is complete for this experiment. Further, to obtain an estimate of occupancy measure, we discretize the two-dimensional state space into $[10,10]$ grid, so that the global occupancy measure for the two-agent problem case is $10^4$ dimensional while the marginalized occupancy measure for each agent is $10^2$ dimensional. 
%
%
%
\begin{figure}[h]
	\centering
	\subfigure[Average return]{\includegraphics[scale=0.25]{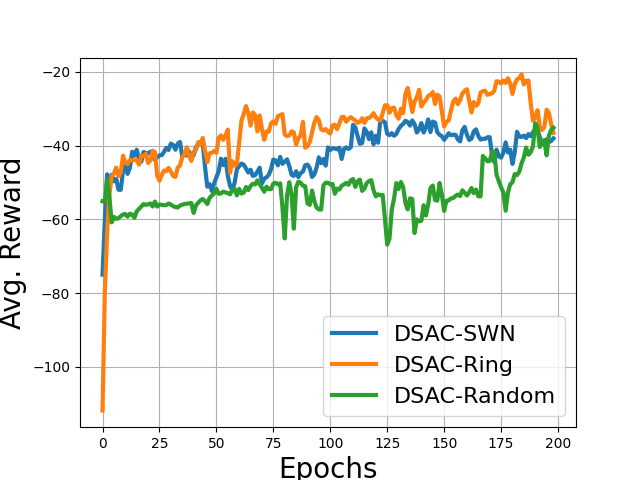}\label{fig_reward_LS1}}
	\subfigure[Average cost]{\includegraphics[scale=0.25]{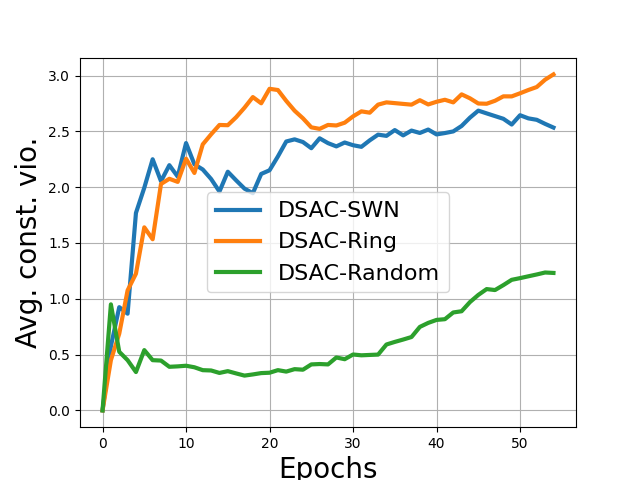}\label{fig_cost_LS1}}
	\hspace{0cm}
	\subfigure[Consensus error]{\includegraphics[scale=0.25]{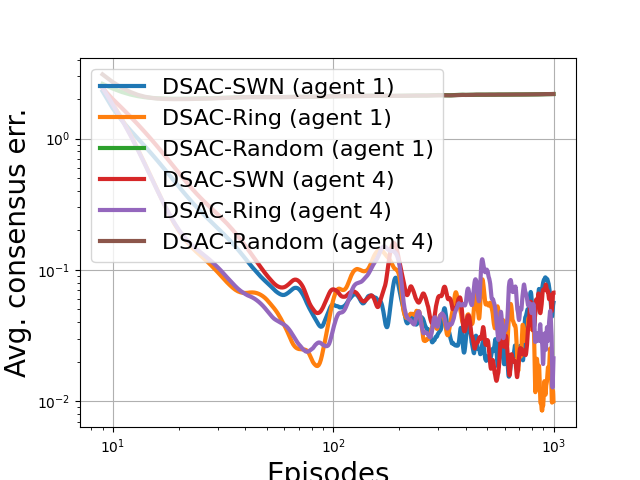}\label{fig_consensus1}}
	\captionof{figure}{Safe navigation in a multi-agent cooperative environment similar to the one mentioned in Fig. \ref{fig_environment_multi} with $8$ agents and $8$ landmarks. Note that the state space in this case would be $16$ dimensional (location of agent and landmarks). We run this experiment for three  communication graphs among agents; \emph{small world network (SWN)} (all the agents are connected by the  repeated generation of
		Watts-Strogatz small-world graphs with parameters $k=3$ and $p=0.5$) , \emph{ring} (all the agents are connected using ring topology), and \emph{random} (where agents are randomly using Erd\H{o}s-R\'enyi random graph model). (a) We plot running average of the reward return similar to Fig. \ref{fig_multi_reward}. (b) We  plot the running average of the constraint violation similar to Fig. \ref{fig_multi_constraint}. (c) We plot the running average of the consensus error for agent $1$ and agent $4$ for \emph{small world network}, \emph{ring}, and \emph{random} network connectivity.}
	\label{fig_LS2}
\end{figure}
\textbf{Safe Cooperative Navigation.} We consider a two agent cooperative environment from \cite{lowe2017multi} where each agent is equipped with the task to reach its assigned goal without traversing via the unsafe region. Note that this behavior could be learned in a policy of a agent $i$ via imposing a safety constraint for each agent $\ip{\lambda^\pi_i,c} \leq C$ where $\lambda^\pi_i$ in the marginalized occupancy measure. This local constraint could be introduced into the global common objective as a quadratic penalty in a manner similar to \eqref{eq:penalty} as 
\begin{equation}\label{eq:penalty_multi}
	F(\lambda^\pi) = \frac{1}{N}\sum_{i=1}^{N}\langle \lambda^\pi_i, r_i\rangle - z\sum_{i=1}^{N} \left(\ip{\lambda^\pi_i,c}-C\right)^2,\vspace{-0.0cm}
\end{equation}
where $z$ is the penalty controlling parameter.
We use the proposed DSAC algorithm to solve problem, and present the results for the average reward and constraint violation, respectively, in Fig. \ref{fig_multi_reward}-\ref{fig_multi_constraint}.  Further, a "demo.gif" file is submitted along with the paper to show the learned policy for $4$ agents. 

To further justify the proposed approach, we consider a set of $8$ agents in a multi-agent cooperative environment similar to the one mentioned in Fig. \ref{fig_environment_multi} but with $8$ agents. Similarly, the actor and critic are parameterized by a two layer DNN with $64$ nodes per layer for each agent. We consider three different scenarios of network connectivity for this experiment namely; \emph{fully connected (FC)} (all the agents are connected to each other), \emph{ring} (all the agents are connected using ring topology), and \emph{random} (where agents are randomly connected using Erd\H{o}s-R\'enyi random graph model with $p$ being uniformly selected between $0$ and $1$). We plot the running average of the reward returns (Fig. \ref{fig_reward_LS1}),  running average of the constraint violations (Fig. \ref{fig_cost_LS1}), and running average of the consensus error (Fig. \ref{fig_consensus1}) for agent $1$ and 4 in Fig. \ref{fig_LS2}. Since the consensus error was converging to zero quickly, we have plotted it using log scale and episodes for the $x$ axis. 
\end{document}